\documentclass{article}

\PassOptionsToPackage{numbers, compress}{natbib}

\usepackage[preprint]{neurips_2023}

\usepackage[utf8]{inputenc} % allow utf-8 input
\usepackage[T1]{fontenc}    % use 8-bit T1 fonts
\usepackage{hyperref}       % hyperlinks
\usepackage{url}            % simple URL typesetting
\usepackage{booktabs}       % professional-quality tables
\usepackage{amsfonts}       % blackboard math symbols
\usepackage{nicefrac}       % compact symbols for 1/2, etc.
\usepackage{microtype}      % microtypography
\usepackage{xcolor}         % colors

\usepackage{amssymb}
\usepackage{amsmath}
\usepackage{booktabs}
\usepackage{graphicx}
\usepackage{caption}
\usepackage{subcaption}

\usepackage{listings}

\makeatletter
\DeclareRobustCommand\onedot{\futurelet\@let@token\@onedot}
\def\@onedot{\ifx\@let@token.\else.\null\fi\xspace}

\makeatother

\newcommand{\mY}{\mathcal{Y}}
\newcommand{\mX}{\mathcal{X}}

\title{Collaborative Blind Image Deblurring}

\author{%
  Thomas Eboli \qquad{} Jean-Michel Morel \qquad{} Gabriele Facciolo\\
  \\
  Universit\'e Paris-Saclay, ENS Paris-Saclay, CNRS, Centre Borelli, France\\
  \\
  \url{https://github.com/teboli/collaborative_blind_deblurring}
}

\begin{document}

\maketitle

\begin{abstract}
Blurry images usually exhibit similar blur at various locations
across the image domain, 
a property
barely captured in nowadays
blind deblurring neural networks. We show that when
extracting patches of similar underlying blur is possible,
jointly processing the stack of patches yields superior
accuracy than handling them separately.
Our collaborative scheme is implemented in a neural architecture with
a pooling layer on the stack dimension.
We present three practical patch extraction strategies for
image sharpening, camera shake removal and optical
aberration correction, and validate the proposed
approach on both synthetic and real-world benchmarks.
For each blur instance, the proposed collaborative strategy
yields significant quantitative and qualitative improvements.
\end{abstract}

%%%%%%%%% BODY TEXT
\section{Introduction}
\label{sec:intro}

Image deblurring is the problem of predicting sharp images from
blurry ones.
In the most common case, a single blurry image
is available, and the goal is to predict its sharp version. Except in rare situations where the blur is known, {\em e.g.,} via
embedded hardware~\cite{park14gyro} for camera shake, or via lens
calibration~\cite{bauer18automatic} for lens blur, 
we have little information on the blur. 
This blind setting is highly ill-posed and requires priors over both the latent sharp image and the blur to be solved~\cite{levin09understanding}.
Classical approaches to blind deblurring leverage image and blur kernel priors to first estimate the blur kernel, and second
estimate a sharp image via non-blind deblurring. The best
priors for doing so, {\em e.g.,}~\cite{pan18dark}, are based on picking the salient edges across the image that provide high-quality
hints on the blur~\cite{zhe15regions}.

Yet, ever since the introduction of large corpora of sharp/blurry image
pairs such as the GoPro dataset~\cite{nah17multiscale}, neural networks
achieve state-of-the-art deblurring results. They are trained to predict a sharp image directly from a blurry input, without requiring the intermediate
estimation of the blur kernel. To do so, the network 
may need to extract some relevant features of 
the blur in order to predict correctly the missing high frequencies. 
The hints needed to determine the blur (the edges~\cite{zhe15regions})
are usually spread across larger regions of the image than what is 
captured by the receptive fields of convolutional networks. This a reason why 
the multi-scale 
architectures~\cite{cho21rethinking, nah17multiscale} are so widespread for this task. Self-attention~\cite{vaswani2017attention} can 
collect such widespread hints, however they are too expensive to 
deploy in realistic scenarios with large images~\cite{liu21swin}.
Despite impressive results, none of these approaches truly leverage the existing image processing expertise. These sparse hints are well exploited by classic methods either by exploiting the global property of the Fourier transform 
~\cite{goldstein12blur} or by analyzing the directional gradient histogram of the image~\cite{delbracio21polyblur}.

In this work we propose to feed the network the relevant hints needed to improve the ability of the model to extract the blur from the representation of the blurry image. We propose a collaborative scheme where the network jointly processes  several patches with the same underlying blur as a way to disambiguate the blur. 
Increasing the number of patches increases the probability of collecting all the useful information needed to deblur.
Collaboration happens by a sort of ``attention-without-attention'' module, which amounts to selecting relevant patches for the task at hand. For instance, in the case of motion deblurring, patches coming from the same moving object. Note that the patches can come from locations thousands of pixels apart, which is in practice hardly achievable by convolutions or attention. 
We show that collecting such sets of patches sharing similar underlying
blur is straightforward for three practical instances of blur: camera shake~\cite{whyte12shake}, optical aberrations~\cite{schuler12blind}
and mild blurs~\cite{delbracio21polyblur}.
Within the network, this collaboration is achieved by processing the patches in parallel and by inserting pooling layers that foster the collaboration between the  encoded features.  
Our experiments show that this collaborative processing boosts the deblurring accuracy of the network. This strategy can be applied to a variety of architectures. A practical
application of our technique illustrated in this paper is designing lightweight
yet efficient blind deblurring networks.

Our contributions are summarized as follows:
(1) We propose a collaborative strategy that consists in gathering a stack of patches
with similar underlying blur in an image, and jointly processing them
    in a neural network upgraded with a layer pooling along the stack dimension. 
(2) We show its practicality for three instances of blur: camera shake, optical aberrations and mild blurs.
(3) We provide theoretical elements to connect the proposed approach
    to existing classical blind deblurring methods.
(4) We show on both real-world and synthetic data the efficiency of the approach for the three sorts of blur listed above, and validate two elements of design: how many patches should collaborate and which pooling function to use.

\section{Related work}
\label{sec:related}

Classic blind image deblurring algorithms alternate between a kernel-estimation step, and a non-blind deblurring step~\cite{levin09understanding}.
At each step the estimated kernel is used to recover a sharper image by  the non-blind deblurring, which in turn is used to refine the kernel prediction.
The kernel prediction makes use of domain knowledge such as 
smoothness and sparsity of camera shake~\cite{whyte12shake},
approximate symmetry of optical aberration~\cite{schuler12blind},
the Gaussian shape of defocus blur~\cite{hasinoff07layerbased}
or translational motions in street photography~\cite{couzinie13learning}.
The work of Nah {\em et al.}~\cite{nah17multiscale} adopts instead a black-box paradigm
by learning from a large dataset of aligned blurry and sharp image pairs
a multi-scale CNN that predicts from a single blurry image a restored variant,
without the need for traditional image priors or any explicit structure
on the family of blur to remove. The properties on the blur are now
determined by the training dataset, for instance camera shake~\cite{rim20dataset}, defocus~\cite{abuolaim20dualpixel} or dynamic motions~\cite{nah17multiscale}.
Subsequent architectures follow this trend by introducing 
recurrent layers~\cite{tao18scale}, additional skip connections~\cite{gao2019dynamic, park20multitemporal}, attention modules~\cite{cho21rethinking, wang22uformer}, adversarial losses~\cite{kupyn19deblurgan}, patch-aware normalization layers~\cite{chu22improving}, and more recently diffusion models~\cite{delbracio2023inversion, whang22refinement}.
Similar deep learning-based strategies have been since proposed for defocus~\cite{abuolaim20dualpixel} and optical aberration~\cite{chen21postprocessing} correction
consisting in both collecting large real-world supervisory datasets and designing ad-hoc models.
In this work, we combine knowledge of a certain kind of blur, {\em e.g.,} camera shake, out-of-focus blur, 
optical aberration, 
and blind deblurring networks
by grouping patches from an input blurry image.
We explicitly make them
interact within an architecture whose design is inspired by the burst deblurring approach of~\cite{aittala18burst}, yet for restoring a {\em single} image. 

Using multiple images or patches from a single image to disambiguate 
restoration is common in image processing. 
Image collaboration
is crucial for system-specific degradation
such as vignetting correction~\cite{lalonde2010sun}, camera calibration~\cite{zhang00calibration} or fixed pattern noise estimation~\cite{chen2008determining}. 
Nevertheless the most notable
example of collaborative image processing is the BM3D~\cite{dabov07bm3d} denoising algorithm 
that gathers patches with similar aspect to denoise them together.
In the same spirit  burst deblurring methods~\cite{aittala18burst,delbracio15fba}
combine multiple frames depicting  the same underlying sharp image but 
different blurs.
Our approach is a single-image method that can 
be seen as the {\em dual} problem of burst deblurring: we select
patches of different underlying sharp contents but sharing identical or similar blurs.
This aims at obtaining information on the blur in as many directions as possible thanks to a greater variety of directional gradients
from the different images/patches
to better predict a blur kernel, a general idea considered in previous works~\cite{delbracio21polyblur, eboli22breaking, goldstein12blur, michaeli14patch}.
To our knowledge this philosophy is explicitly applied to single-image
blind deblurring via neural networks for the first time in this paper.

\begin{figure}[t]
    \centering
    \begin{tabular}{cc}
        \begin{subfigure}[t]{0.48\linewidth}
            \centering
            \includegraphics[width=\linewidth,trim=10 15 10 10,clip]{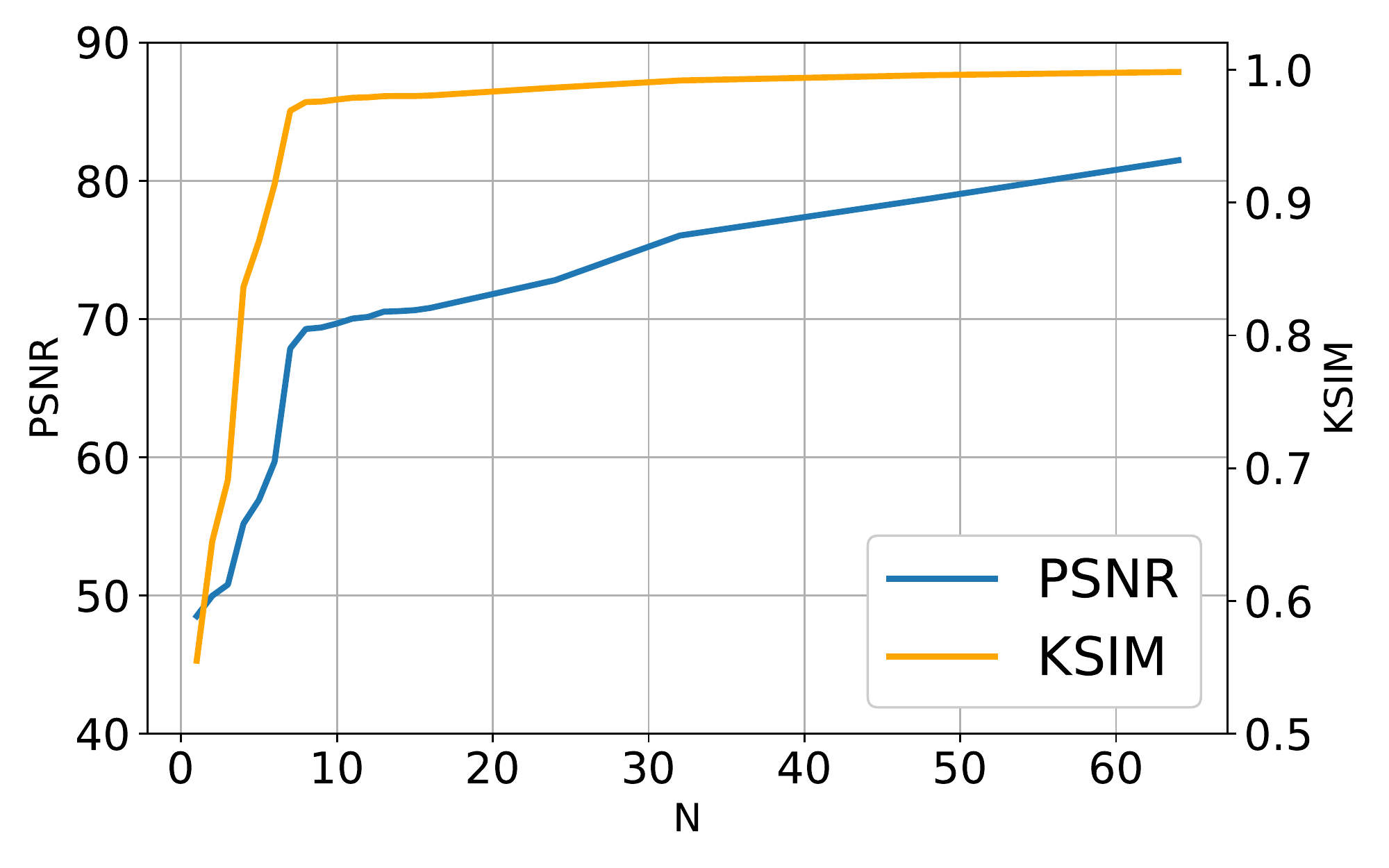}
    \caption{Average of the 8 kernels from~\cite{levin09understanding}.}
        \end{subfigure} &
        \begin{subfigure}[t]{0.48\linewidth}
            \centering
            \includegraphics[width=\linewidth,trim=50 50 50 42,clip]{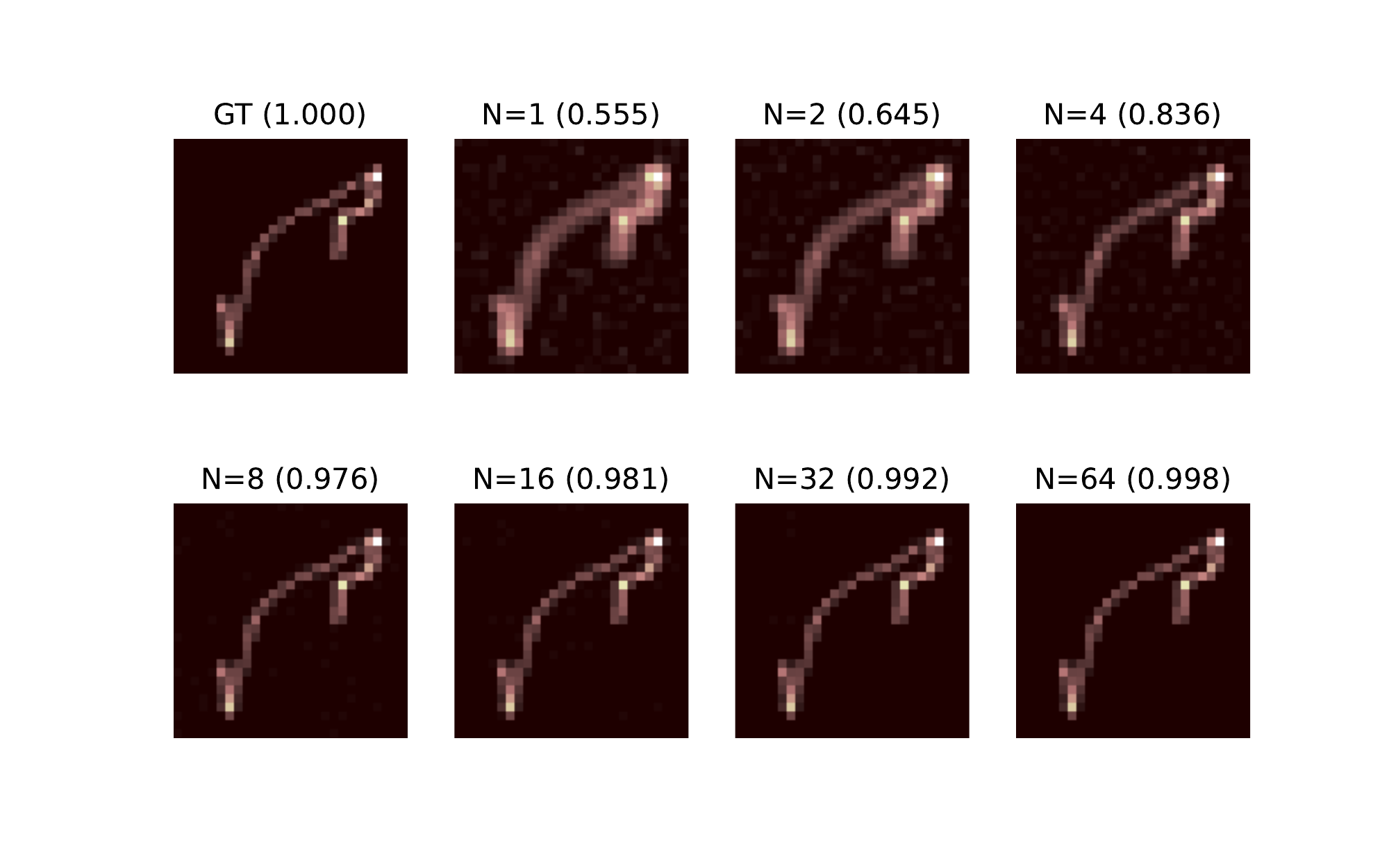}
    \caption{The fourth kernel from~\cite{levin09understanding}.}
        \end{subfigure} \\
    \end{tabular} 
    \caption{Illustration of the collaborative scheme with the kernels from \cite{levin09understanding}. We report the PSNR and the kernel similarity (KSIM) as defined in \cite{li13unnatural} on the left
    and the KSIM for each qualitative kernel on the right. Combining more sharp/blur pairs dramatically improves
    the accuracy of the kernel support, with saturation occuring at $N=8$.}
    \label{fig:kernels}
\end{figure}

\section{Theoretical background}
\label{sec:background}

To motivate our approach we review the problem of kernel estimation from several samples.
Let $x_1, x_2, \dots, x_N$ be $N$ sharp images, 
$y_1, y_2, \dots, y_N$ corresponding blurry images  where
$y_n = k \ast x_n + \varepsilon_n$
for $n$ in $\{1,\dots,N\}$, $k$ a blur kernel, and the $\varepsilon_n$'s instances of 
noise.
From the pairs $(x_n, y_n)$ ($n=1,\dots,N$), we estimate the blur kernel
$k$ by minimizing a regularized $\ell_2$ energy function:
\begin{equation}
    \min_k \frac{1}{N}\sum_{n=1}^N\|y_n - k\ast x_n\|^2_2 + \lambda \|k \|^2_2.
\end{equation}
Its unique minimizer may be obtained in the Fourier domain by
\begin{equation}\label{eq:div}
    \widehat{K} = \sum_{n=1}^N X_n^\ast Y_n / \left(\sum_{n=1}^N X_n^\ast X_n + \lambda N\right),
\end{equation}
where the capital letters $X_n$, $Y_n$ and $K$ denote the Fourier transforms
of $x_n$, $y_n$ and $k$, the $\ast$ superscript denotes the complex conjugate, and $\lambda$ is a positive regularization weight. 
In the Fourier domain, the multiplication and division are
entrywise.
In this setting, even if one image $X_n$ has a zero at a given frequency, the average of several images would likely not be zero, unless the frequency is removed by the blur kernel. When $N=1$
instead, if there is ambiguity at some frequencies (the blur may be oriented in the same direction as an edge for instance), then
it is nearly impossible to recover the correct frequency of the blur
kernel.
On the spatial domain, considering more images thus
boils down to gathering as much information on the oriented
gradients as possible to disambiguate the blur from
the signal, a strategy at the core of certain blind deblurring techniques~\cite{delbracio21polyblur, goldstein12blur}.

Let us illustrate how important it is to use $N$ image pairs
{\em together} using Eq.~\eqref{eq:div}.
We select the $512\times512$ central crops of the 64 first
RGB test images from the DIV2K dataset~\cite{agustsson17div2k} 
as sharp images $x_n$ $(n=1,
\dots,64)$ in order to have all the images at the same format. 
We blur them with the 8 kernels from
the Levin dataset~\cite{levin09understanding}, and 
add $1\%$ white Gaussian noise.
Figure~\ref{fig:kernels} illustrates the increase of the accuracy
of the kernel support with respect
to $N$ for the 8 kernels from \cite{levin09understanding}.
The results confirm that as the number of images in the stack increases, the performance improves significantly, leading to a nearly perfect reconstruction of even challenging a motion blur kernel. Notably, the performance saturates at a relatively small stack size of $N=8$.
This analysis highlights the dramatic improvement that collaboration
among images with the same blur can bring towards better deblurring.

\section{Proposed method}
\label{sec:method}

In this section we address the more realistic blind deblurring case, where only the blurry images $y_1, \cdots y_N$ are available.  We thus cannot rely directly on Eq~\eqref{eq:div} to improve the kernel estimation. 
We instead leverage the collaboration of $N$ blurry images, as highlighted in the previous section, in order to better capture the structure of the blur while at the same time deblurring. This collaborative processing results in an efficient single-image blind deblurring neural network.

\subsection{Collaborative architecture}
\label{subsec:architecture}

We propose a neural network $f_\theta$ with parameter $\theta$
that features two notable changes compared to
typical single-image blind deblurring networks, {\em e.g.,} \cite{nah17multiscale}: (i) $N$ inputs and outputs instead of a single one, and (ii)
an inner collaboration layer combining the $N$ feature
maps.
When $N$ is 1, the proposed framework
boils down to the classical single-image blind deblurring approach.

\paragraph{Input and output.}
We propose a neural network that processes in a single forward pass $N$ images
$\mY_N = \{y_1, \dots, y_N\}$ containing similar blurs 
(not necessarily the exact same for each image), 
and predicts $N$ sharp versions
$\widehat{\mX}_N = \{\widehat{x}_1, \dots, \widehat{x}_N\}$.
In practice these sets are implemented as 4D tensors
concatenating all the RGB images of the stacks, the first 
dimension being of size $N$.
Let $f$ be such a network and $\theta$ its parameter, then the inference
reads
\begin{equation}\label{eq:cnn}
    \widehat{\mX}_N = f_\theta\left(\mY_N\right).
\end{equation}
During training, the $N$ restored patches have collaborated
and can thus be supervised individually since what matters in the
end is the quality of each individual image $\widehat{x}_n$ ($n=1,\dots,N$).
Provided training pairs of sets $\mX_N^{(m)}$ and $\mY_N^{(m)}$ $(m=1,\dots,M)$,
we learn the parameter $\theta$ by minimizing:
\begin{equation}\label{eq:loss}
    \min_\theta \frac{1}{MN}\sum_{m=1}^M\sum_{n=1}^N \ell(\widehat{x}_n^{(m)}, y_n^{(m)}),
\end{equation}
where $\ell$ is a pixelwise loss. In this work, we adopt
the $\ell_1$ distance as supervising loss function.
The images $\widehat{x}_n^{(m)}$ and $y_n^{(m)}$ are the $n$-th
elements of respectively $\widehat{\mX}_N^{(m)}$ and $\mY_N^{(m)}$.

\paragraph{Inner collaboration.}
We follow the approach of Aittala and Durand~\cite{aittala18burst}. This approach applies the same convolution layers to each one of the $N$ images of the stack
and combine the knowledge from different images in the same stack
via pooling of the individual blurry images' representations after 
a given convolutional or attention layer in a network.
Let $e_n$ be the representation of the $n$-th blurry 
image $y_n$ ($n=1,\dots,N$) after this given layer of $f_\theta$. We implement the inner collaboration
layer by a pooling function $p$ operating on the $e_n$ along 
the stack dimension, and that returns a global 
representation $g$ of the stack:
\begin{equation}\label{eq:pooling}
    g = p(\{e_1, \dots, e_N\}).
\end{equation}
The feature $g$ has the same dimension as an individual local feature $e_n$ ($n=1,\dots,N$).
In \cite{aittala18burst}, in the context
of burst deblurring, $g$ is supposed
to extract a representation of the underlying sharp signal since
all the blurry frames in the burst have different blurs but
share the same underlying sharp content.
In our case the roles of the images and blurs are reversed: we have 
images of different contents sharing similar blurs in our stack. Consequently,
we expect  the global information compiled in $g$ via 
the pooling function $p$ to be related to the blur.
Lastly, the local features $e_n$ ($n=1,\dots,N$)
are updated through the merge with the global representation $g$
via a $1\times1$ convolution layer:
\begin{equation}\label{eq:conv1x1}
    e_n \gets \texttt{conv}_{1\times1}(\{e_n, g\}), \quad\forall n \in \{1,\dots,N\}.
\end{equation}
That way, we expect the upgraded feature map $e_n$ to be guided
by some information on the blur, ultimately improving the
deblurring ability of $f_\theta$.

The function $p$ has not been explicitly defined so far on purpose.
Such a function should take a stack of spatial feature maps
and return a single spatial feature map. Since no
assumption is made on $p$, it can be either learning-free
or learnable.
Classical pooling strategies such as the mean or max
functions are used in \cite{aittala18burst} and are
drop-in candidates to extract $g$ in our context.
We also explore in this paper how to learn $p$
by implementing it with the lambda layer from \cite{bello21lambda}
or the self-attention (SA) module~\cite{vaswani2017attention}.
The learning-free approach is important to build small
models that may be deployed on devices like smartphones~\cite{ma22searching}, whereas 
the learnable modules aim at better performance at the
cost of additional computations.
We benchmark the different candidates for $p$ in Section~\ref{sec:experiments}.

\begin{figure}[t]
    \centering
    \resizebox{\linewidth}{!}{ 
    \begin{tabular}{ccc}
        \begin{subfigure}[t]{0.33\textwidth}
            \centering
            \includegraphics[width=\textwidth]{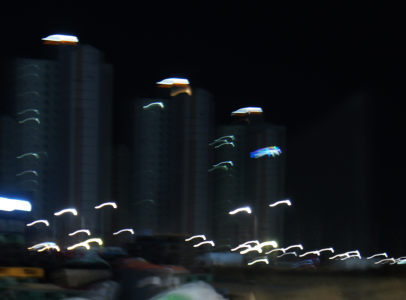}
            \caption{Camera shake~\cite{rim20dataset}.}
        \end{subfigure} &
        \begin{subfigure}[t]{0.33\textwidth}
            \centering
            \includegraphics[width=\textwidth]{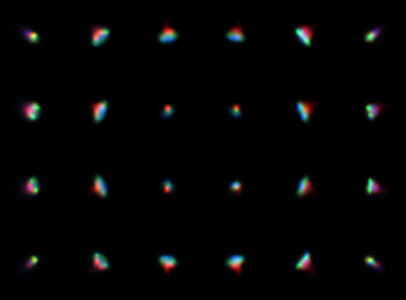}
            \caption{Optical aberration~\cite{bauer18automatic}.}
        \end{subfigure} &
        \begin{subfigure}[t]{0.33\textwidth}
            \centering
            \includegraphics[width=\textwidth]{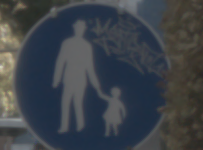}
            \caption{Fusion blur~\cite{lafenetre23handheld}.}
        \end{subfigure} \\
    \end{tabular} 
    }
    \caption{Three real instances of blur: camera shake from the RealBlur dataset~\cite{rim20dataset} (scene 118, image 2), a subset of the local calibrated optical aberrations from the Canon EF24mm f/1.4L USM opened at f/2.8, calibrated in \cite{bauer18automatic}, and the result
    of the $\times2$ multi-frame super-resolution online demo of \cite{lafenetre23handheld}. For shake the light
    streaks suggest the blur is roughly the same everywhere, thus patches can be sampled uniformly. The same holds for fusion blur where the fused image globally lacks of sharpness. The aberrations are all unique but roughly follow a central symmetry, thus easy to sample.}
    \label{fig:samekernels}
\end{figure}

\subsection{Gathering images with similar underlying blurs}
\label{subsec:inputs}

The theoretical background of the previous section and  its proposed integration within a neural network
assume that $N$ images $y_n$ ($n=1,\dots,N$) degraded with similar 
blurs are given. For camera shake~\cite{whyte12shake}, mild blurs~\cite{delbracio21polyblur} (covering out-of-focus and sharpening),
and optical aberrations~\cite{eboli22fast}, such patches
may be handpicked from a {\em single} image.
For camera shake the blur smoothly varies across the field of view~\cite{whyte12shake}, and
for ``reasonable'' shake like small translations featured in the Realblur
dataset~\cite{rim20dataset}, it boils down to a uniform blur kernel applied to the whole image: any
crop in the image therefore has the same blur.
Rough uniformity of the blur is also a typical assumption for modeling
lack of sharpness in the image, {\em e.g.,} because of a failing
autofocus or the interpolation blur of a preceding multi-frame algorithm~\cite{lecouat22hdr}.

For instance, inevitable blur may come from optical aberrations~\cite{eboli22breaking, kee11modeling}. 
Very similar blurs may be found at symmetric
locations on the field of 
view~\cite{schuler12blind}.
Sharpening is another instance of important brick in 
most ISP pipelines nowadays~\cite{delbracio21polyblur}, 
and consists in removing small Gaussian-like blurs
caused by slight out-of-focus or multi-frame fusion algorithms.
The blur may vary but it may be considered roughly similar
across the field of view, for instance to remove the fusion blur
introduced by multi-frame algorithms~\cite{eboli22breaking}.
A third common category is camera shake~\cite{whyte12shake}
during exposure, resulting in global motion blur across
the image. A single blur kernel~\cite{rim20dataset} or smoothly varying blurs~\cite{whyte12shake} may model the whole
shake, thus validating the assumption of similar kernels across the image. 
An illustration is shown in Figure~\ref{fig:samekernels}.
In the three common sorts of blur list above, grouping 
patches with similar blur is thus feasible by leveraging
the properties of the blur.
Collecting patches that way amounts to finding the most relevant patches for a given deblurring task, which is a sort of handcrafted attention or ``attention-without-attention''. Grouping in that manner patches, which may be hundreds or thousands of pixels away from each other,
is indeed doable with CNNs or attention, but at the cost of either very deep models or important computation, all that for computing something that could be known in advance.

Note that we have not discussed  defocus blur
where local kernels of similar aspect might be easily
grouped if the depth of the scene is known~\cite{hasinoff07layerbased},
for instance with recent monocular
depth estimators, {\em e.g.,} \cite{ranftl21transformer}.
Nevertheless, we keep this multimodal approach for
future work and focus instead on sorts of blurs where the grouping can be done manually as for the examples in Figure~\ref{fig:samekernels}.

\section{Experiments}
\label{sec:experiments}

Experiments were all run on a single 16Gb NVIDIA V100
graphic card. Descriptions of the training and evaluation protocols, and qualitative results are in the supplementary material.

\subsection{Collaborative model}
\label{subsec:model}

In this work, we use a UNet model that is a general architecture 
used as the foundation of many practical deblurring models, 
{\em e.g.,}~\cite{chen21postprocessing, lai22face}.
Since it is an all-purpose model, it has no specific bias for removing
blur in contrast to the state-of-the-art CNNs~\cite{cho21rethinking,tao18scale}.
It is thus an adequate model to measure the impact of the proposed
collaborative scheme.
We introduce two variants called UNet and UNet-T (for tiny)
embedding respectively 4 and 3 downsampling/upsamling layers
in the encoder/decoder: The initial number of feature maps $C$ is 64
and the respective bottleneck sizes of these models are 512 and 256 channels. 
These models have respectively around 17M and 4.3M parameters.
Before each down/upsampling
module we place $p$ to enforce
collaboration. 

So far we have only presented the broad idea that features
should be shared within $p$, but we did
not delve into details. We compare the max pooling approach of
\cite{aittala18burst} for burst deblurring,
the lambda layer~\cite{bello21lambda} and a self-attention (SA)
layer as in the Transformer architecture \cite{vaswani2017attention}.
The lambda layer is implemented with 4 feature channels. The three-layer 
perceptron of the SA layer is shaped as an inverted bottleneck.
We also evaluated the feature mean to implement $p$ but
observed as in~\cite{aittala18burst} that it leads
to the same results as the max pooling.
Diagrams are in the supplementary material.

\subsection{Validation on Gaussian blur}
\label{subsec:ablation}

We start with 2D anisotropic Gaussian blur kernels that may approximate
several instances of real-world blur such as lens blur~\cite{kee11modeling}, 
defocus~\cite{hasinoff07layerbased}, and translational motion blur~\cite{delbracio21polyblur}.
Evaluating our approach on Gaussian blur is thus a simple manner to
validate our approach in a controlled setting that corresponds
to many real-world blurring scenarios.
We train UNet and UNet-T on $128\times128$ crops randomly sampled
from the 800 training images of the DIV2K dataset~\cite{agustsson17div2k}.
We randomly flip and rotate the patches prior to blurring them
with Gaussian blur kernels of standard deviation along the two principal axes 
uniformly sampled in $[0.3,4]^2$, {\em i.e.,} up to a $33\times33$ blur spot,
and the orientation is uniformly sampled in $[0,2\pi)$ (same model as in \cite{delbracio21polyblur}).
We add moderate Gaussian noise with standard deviation 
randomly sampled in $[0.5,2]/255$ after blurring.

\begin{minipage}{0.5\textwidth}
  \centering
  \begin{tabular}{cll}
    \toprule
     $N$ & UNet & UNet-T \\
    \midrule
    $1$  & 32.75 & 32.46 {\scriptsize (-0.29)}\\
    $2$  & 32.91 & 32.75 {\scriptsize (-0.16)}\\
    $4$  & 32.92 & 32.92 {\scriptsize (+0.00)}\\
    $8$  & 33.12 & 33.13 {\scriptsize (+0.01)}\\
    $16$ & 33.02 & 33.07 {\scriptsize (+0.05)}\\
    \bottomrule
  \end{tabular}
  \captionof{table}{Average PSNR over 400 images blurred with isotropic Gaussian blur. UNet-T achieves the same performance as UNet for $N\geq4$ despite having $4$ times more
  parameters.}
  \label{tab:ablationdepth}
\end{minipage}%
\begin{minipage}{0.5\textwidth}
  \centering
 \includegraphics[width=\textwidth]{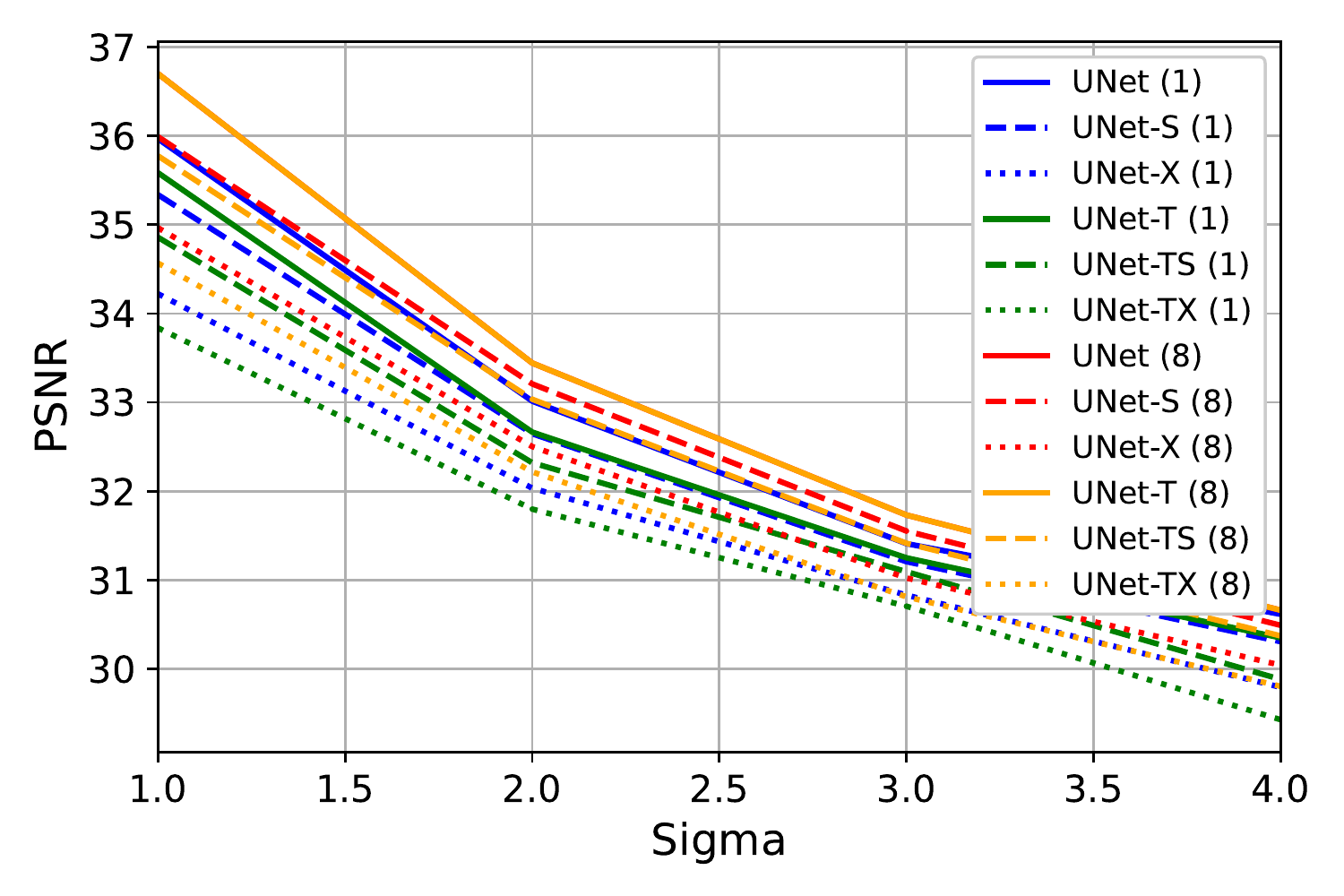}
  \captionof{figure}{PSNR of UNet and UNet-T for 3 widths. $N$ is in parenthesis. For each width, the models with $N=8$ are above.
  The plain red curve is covered by the orange one.}
  \label{fig:ablationwidth}
\end{minipage}

\paragraph{Image stack size $N$.} 
In order to evaluate the choice of the
stack size $N$, we train networks with input stacks of
blurry images sharing the same blur with a unique stack size $N$ in
$\{1,2,4,8,16\}$. We choose the max pooling strategy to not introduce
additional learnable parameters, and isolate the impact
of using several patches together.
We train the models for 250k iterations with the Adam optimizer
and initial learning rate of $5\times10^{-5}$ that is reduced 
by 0.5 after each 40k iteration until reaching $10^{-6}$ 
where we observe convergence.
We evaluate the models on sets of 100 images blurred with an anisotropic
Gaussian blur kernel with standard deviation $\sigma$ in $\{1,2,3,4\}$,
and with additional $0.5/255$ Gaussian noise.
We show in Table~\ref{tab:ablationdepth} that the performance of both models grows
with $N$. For $N \geq 4$, the performance of both models is similar
despite UNet having $4$ times more parameters than UNet-T.
All the results are averaged over
3 shuffles of the test images to  draw different images per $N$-sized stack across different runs, but the error bars are marginal and thus not reported.

\paragraph{Reducing UNet.}
Since using $N$ images instead of 1 actually compensates the difference of depth and parameters between UNet and UNet-T for $N\geq 4$, we also verify that
by reducing the width of the UNet, using $N=8$ images
helps to maintain good performance.
Besides training UNet (resp. UNet-T) with the original number of channels
per feature like in the previous experiment, we propose 
the Slim and Extra-Slim variants dubbed UNet-S and UNet-X (resp. UNet-TS and UNet-TX) where the number of
channels for each feature map is respectively divided by 2 and 4, {\em i.e.,}
the bottleneck layer of UNet initially having 512 channels has now
respectively 256 and 128 channels. 
For $N=1$, these models have respectively 4.4M, 1.1M, 1.1M and 270K parameters, and about 5\% more for $N>1$ (because of the 
$1\times1$ convolution layers in Eq.~\eqref{eq:conv1x1}).
We train the four new variants of
UNet and UNet-T with the same procedure, evaluate them on the same 400 test images
as previously, and show the average PSNR per value of $\sigma$ 
in Figure~\ref{fig:ablationwidth}.
It can be seen that the models with $N=8$ systematically achieve better results
than their $N=1$ counterparts. Remark that UNet-TX for $N=8$ achieves results
similar to models that have $\times4$ more parameters, and is between +0.3 and +0.8dB above
its $N=1$ counterpart.
This observation suggests that the learning-free max pooling operation helps
to design lightweight blind deblurring networks with performance on par with larger ones.
UNet-TX has a number of parameters comparable to that of the lightweight demosaicking
networks proposed in \cite{ma22searching}, which is akin to edge computing. This suggests that UNet-TX may be a valid candidate to replace
mild blur deblurring algorithms, {\em e.g.,}~\cite{delbracio21polyblur, eboli22breaking},
in terms of parameters and thus energy consumption.
More quantitative results are in the supplementary material.

\begin{figure}[t]
    \centering
    \resizebox{\linewidth}{!}{
    \begin{tabular}{cccccc}
        \begin{subfigure}[t]{0.17\linewidth}
            \centering
            \includegraphics[width=\linewidth]{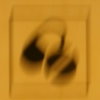}
            \caption{Blurry.}
        \end{subfigure} &
        \begin{subfigure}[t]{0.17\linewidth}
            \centering
            \includegraphics[width=\linewidth]{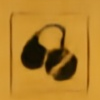}
            \caption{UNet-T.}
        \end{subfigure} &
        \begin{subfigure}[t]{0.17\linewidth}
            \centering
            \includegraphics[width=\linewidth]{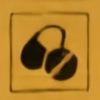}
            \caption{UNet-T (A).}
        \end{subfigure} &
        \begin{subfigure}[t]{0.17\linewidth}
            \centering
            \includegraphics[width=\linewidth]{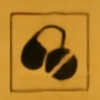}
            \caption{UNet (A).}
        \end{subfigure} &
        \begin{subfigure}[t]{0.17\linewidth}
            \centering
            \includegraphics[width=\linewidth]{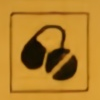}
            \caption{MIMOUNet.}
        \end{subfigure} &
        \begin{subfigure}[t]{0.17\linewidth}
            \centering
            \includegraphics[width=\linewidth]{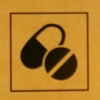}
            \caption{Target.}
        \end{subfigure} \\
    \end{tabular} 
    }
    \caption{An example from Realblur-J~\cite{rim20dataset} deblurred with
    UNet-T, the versions of UNet and UNet-T embedding attention with $N=16$ and MIMOUNet~\cite{cho21rethinking}. The collaboration improves the deblurring accuracy of UNet, to match that of MIMOUNet.}
    \label{fig:shake}
\end{figure}

\subsection{Practical applications}

We illustrate the method on the three practical problems
listed in Figure~\ref{fig:samekernels}:
optical aberration correction, camera shake compensation, and
image sharpening.

\begin{table}[t]
    \centering
    \begin{tabular}{llllc}
    \toprule
        Method & PSNR & SSIM & LPIPS & Params \\
    \midrule
        DeblurGAN-v2$^\ast$~\cite{kupyn19deblurgan} & 29.69 & 0.870 & - & - \\
        SRN$^\ast$~\cite{tao18scale} & 31.38 & 0.909 & - & 6.8M\\
        MIMOUNet~\cite{cho21rethinking} & 30.32 & 0.897 & 0.097 & 6.8M \\
        \midrule
        UNet-T ($N=1$) & 29.30 & 0.870 & 0.133 & 4.3M\\
        UNet-T ($N=4$, M) & 29.88 {\scriptsize (+0.58)} & 0.883 {\scriptsize (+0.013)} & 0.121 {\scriptsize (-0.012)} & 4.6M \\
        UNet-T ($N=16$, M) & 30.33 {\scriptsize (+1.03)} & 0.893 {\scriptsize (+0.023)} & 0.108 {\scriptsize (-0.025)} & 4.6M \\
        UNet-T ($N=4$, L) & 29.96 {\scriptsize (+0.66)} & 0.885 {\scriptsize (+0.015)} & 0.117 {\scriptsize (-0.016)} & 5.6M \\
        UNet-T ($N=16$, L) & \underline{30.45} {\scriptsize (+1.15)} & \underline{0.896} {\scriptsize (+0.026)} & \underline{0.104} {\scriptsize (-0.029)} & 5.6M \\
        UNet-T ($N=4$, A) & 30.13 {\scriptsize (+0.83)} & 0.888 {\scriptsize (+0.028)} & 0.111 {\scriptsize (-0.022)} & 6.4M \\
        UNet-T ($N=16$, A) & \textbf{30.63} {\scriptsize (+1.33)} & \textbf{0.900} {\scriptsize (+0.030)} & \textbf{0.098} {\scriptsize (-0.035)} & 6.4M \\
        \midrule
        UNet ($N=1$) & 30.16 & 0.893 & 0.104 & 17.7M\\
        UNet ($N=4$, M) & 30.45 {\scriptsize (+0.29)} & 0.897 {\scriptsize (+0.004)} & 0.104 {\scriptsize (-0.000)} & 18.7M \\
        UNet ($N=16$, M) & 30.62 {\scriptsize (+0.46)} & 0.900 {\scriptsize (+0.007)} & 0.099 {\scriptsize (-0.005)} & 18.7M \\
        UNet ($N=4$, L) & 30.61 {\scriptsize (+0.45)} & 0.900 {\scriptsize (+0.007)} & 0.098  {\scriptsize (-0.006)} & 22.9M \\
        UNet ($N=16$, L) & \underline{30.78} {\scriptsize (+0.62)} & \underline{0.904} {\scriptsize (+0.011)} & \underline{0.094} {\scriptsize (-0.010)} & 22.9M \\
        UNet ($N=4$, A) & \underline{30.73} {\scriptsize (+0.57)} & \underline{0.904} {\scriptsize (+0.011)} & \underline{0.092} {\scriptsize (-0.012)} & 26.4M \\
        UNet ($N=16$, A) & \textbf{30.98} {\scriptsize (+0.82)} & \textbf{0.908} {\scriptsize (+0.015)} & \textbf{0.087} {\scriptsize (-0.017)} & 26.4M \\
    \bottomrule
    \end{tabular}
    \caption{Results on Realblur-J~\cite{rim20dataset}. The models with $^\ast$ are those reported in \cite{rim20dataset} and are used as setters of the expected ballpark of metrics. MIMOUNet~\cite{cho21rethinking} is retrained following our protocol. Bold and underlined numbers indicate which variants of UNet are the best and second-to-best per UNet model. The difference with the corresponding UNet(-T) with $N=1$ is shown in parenthesis.}
    \label{tab:motion}
\end{table}

\paragraph{Camera shake.}
An important source of blur in personal photography (especially with
handheld smartphones) is camera shake.
We train on the RealBlur-J dataset~\cite{rim20dataset}, composed 3760 sharp/blurry
image pairs of static scenes taken with complex natural camera motions, a UNet with and without
the collaborative layer, and the MIMOUNet~\cite{cho21rethinking} specifically
designed for this task.
Our goal here is not to propose  new state-of-the-art motion blur
models, but rather use the RealBlur dataset as a real-world quantitative
benchmark to quantitatively measure the impact of the proposed collaborative scheme.
We show in Table~\ref{tab:motion} qualitative results on the 
test sets of the RealBlur datasets composed of 980 blurry/sharp image pairs. 
We run UNet and UNet-T by splitting $512\times512$ random crops from the 
training set into a unique stack of size $N$ in
$\{1,4,16\}$ with $25\%$ of overlap to take into account the patches will be stitched
back into a full-sized image for evaluation with the protocol of \cite{rim20dataset}.
During training, we supervise each patch of the stack
with the corresponding patch in the target with 
the loss \eqref{eq:loss}.
At test time, we start from the full image, slice it into $N$ patches, and stitch together the predicted sharps variants into a full sharp estimate with the same windowing
approach as in \cite{schuler12blind}.
We benchmark different choices of pooling layer $p$: max (M),
lambda (L) and SA (A).
We use the same training strategy as MIMOUNet and also
retrain the latter for fair comparison: We train for
1k epochs with the Adam optimizer with initial
learning rate set to 0.0001 and decayed by 0.5 every 200
epoch. The batch size is set to 8.
We see that increasing the stack size $N$ within the 
{\em same} patch and more refined pooling $p$ benefit the deblurring accuracy
for all metrics when using a UNet not initially designed
for deblurring, in contrast to \cite{tao18scale} and \cite{cho21rethinking}. 
For instance Remark a 17M-parameter Unet cannot beat MIMOUNet~\cite{cho21rethinking}, but after being upgraded,
the performances are boosted by at margins up to +1.33dB.
UNet-T with $p$ implemented with SA and $N=16$ notably cope
with MIMONet with less parameters. We also observe that both
refining $p$ and increasing $N$ increase the performance of UNet,
validating our strategy. 
An example is shown in Figure~\ref{fig:shake}.

\begin{table}[t]
    \centering
    \begin{tabular}{llllc}
        \toprule
        Location & Method & 16mm f/2.8 & 24mm f/1.4 & Params \\
        \midrule
        Corner & UNet-TX ($N=1$) &  30.58 {\scriptsize (-0.44)} & 32.83 {\scriptsize (-0.58)} & 270K \\
        Corner & UNet-TX ($N=4$, M) & 31.02   & 33.45 & 284K \\
        Corner & UNet-TS ($N=1$) & 31.60 {\scriptsize (-0.22)} & 34.10 {\scriptsize (-0.47)} & 1.1M \\
        Corner & UNet-TS ($N=4$, M) & 31.82   & 34.57 & 1.1M \\
        Corner & UNet-T ($N=1$) &  \textbf{32.21} {\scriptsize (-0.06)} & 34.88 {\scriptsize (-0.35)} & 4.3M \\
        Corner & UNet-T ($N=4$, M) &  \textbf{32.27} & \textbf{35.32} & 4.6M \\
        \midrule
        Intermediate & UNet-TX ($N=1$) & 34.45 {\scriptsize (-0.34)} & 34.89 {\scriptsize (-0.14)} & 270K \\
        Intermediate & UNet-TX ($N=4$, M) &  34.79 & 35.03 & 284K \\
        Intermediate & UNet-TS ($N=1$) &  35.12 {\scriptsize (-0.37)} & 35.55 {\scriptsize (-0.18)} & 1.1M \\
        Intermediate & UNet-TS ($N=4$, M) & 35.49 & 35.73 & 1.1M \\
        Intermediate & UNet-T ($N=1$) & 35.89 {\scriptsize (-0.23)} & \textbf{36.22} {\scriptsize (-0.01)} & 4.3M \\
        Intermediate & UNet-T ($N=4$, M) & \textbf{36.12} & \textbf{36.23} & 4.6M \\
        \midrule
        Center & UNet-TX ($N=1$) & 36.94 {\scriptsize (-0.57)} & 36.71 {\scriptsize (-0.21)} & 270K \\
        Center & UNet-TX ($N=4$, M) & 37.57 & 36.92 & 284K \\
        Center & UNet-TS ($N=1$) &  38.02 {\scriptsize (-0.36)} & 37.93 {\scriptsize (-0.03)} & 1.1M \\
        Center & UNet-TS ($N=4$, M) & 38.38  & 37.96 & 1.1M \\
        Center & UNet-T ($N=1$) & 39.10 {\scriptsize (-0.18)} & \textbf{38.99} {\scriptsize (-0.00)} & 4.3M \\
        Center & UNet-T ($N=4$, M) & \textbf{39.28} & \textbf{38.99} & 4.6M \\
        \bottomrule
    \end{tabular}
    \caption{Optical aberration removal for
    two lenses calibrated in \cite{bauer18automatic}. The PSNR is reported from
    three different locations on the field of view. The difference between using 1 or 4 images is shown in parenthesis. Setting $N$ to 4 helps compensating the reduction of the width and parameters.}
    \label{tab:aberrations}
\end{table}

\paragraph{Optical aberrations.}

We leverage the near central symmetry of aberrations around
the optical center of the lens~\cite{schuler12blind} by sampling
at $N=4$ locations on the field of view, one for each quadrant of the
Cartesian plane, (see Figure~\ref{fig:samekernels}).
Since no real-world pairs of aberrated and aberration-free image dataset
exists, we generate synthetic data from two of the 70 PSFs calibrated in \cite{bauer18automatic}. Each PSF consists of about 4,000 local RGB
kernels accounting for both monochromatic and chromatic aberrations.
We select the Canon EF16-35mm f/2.8L USM EI at shortest focal length
and maximal aperture and the Canon EF24mm f/1.4L USM that are prone to aberrations after visual inspection, the former being poorer than the latter.
We convert from JPEG to pseudo-linear RGB images the $128\times128$ 
patches from the DIV2K images with the protocol of \cite{brooks19unprocessing},
and blur each color channel with the corresponding one from
local filters sampled at random in the given PSF. We add
$0.5/255$ Gaussian noise to account for noise residual after
demosaicking in an ISP pipeline. 
We train for each PSF , UNet-T, UNet-TS and UNet-TX 
for $N=1$ and $N=4$ with max pooling
as collaborative strategy since in an ISP pipeline within a handheld camera,
each module should be as lightweight as possible.
We train for 500k iterations with the Adam optimizer with batch size of 16,
and an initial learning rate of $10^{-4}$ decayed by 0.2 after
200k and 400k iterations.
Table~\ref{tab:aberrations} shows the PSNR for three locations on the lens field-of-view with growing loss of quality from
the center to a corner. For both lenses the collaborative variant is always
above the model with $N=1$ with significant margins between +0.2 and +0.6dB for UNet-TX. We observe more important improvements for the 16mm lens than the 24mm. Since the
former is a zoom lens, it has in average a poorer quality than a prime lens, thus
benefiting more from collaboration, especially for UNet-TX, validating
the assumption that collaboration helps to design practical
lightweight networks. We can reasonably expect that on
devices with worse optical quality like smartphones, collaboration may help even
further.

\begin{figure}[t]
    \centering
    \resizebox{\linewidth}{!}{
    \begin{tabular}{cccc}
        \begin{subfigure}[t]{0.25\linewidth}
            \centering
            \includegraphics[width=\linewidth,trim=10 10 10 10,clip]{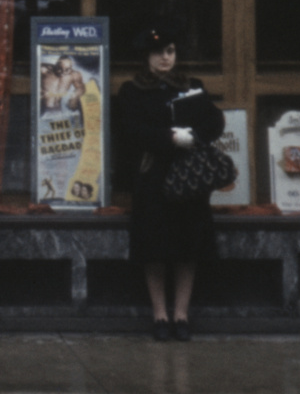}
            \caption{Blurry.}
        \end{subfigure} &
        \begin{subfigure}[t]{0.25\linewidth}
            \centering
            \includegraphics[width=\linewidth,trim=10 10 10 10,clip]{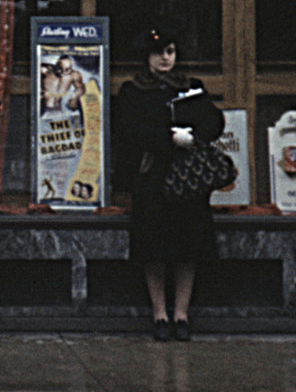}
            \caption{Unsharp mask.}
        \end{subfigure} &
        \begin{subfigure}[t]{0.25\linewidth}
            \centering
            \includegraphics[width=\linewidth,trim=10 10 10 10,clip]{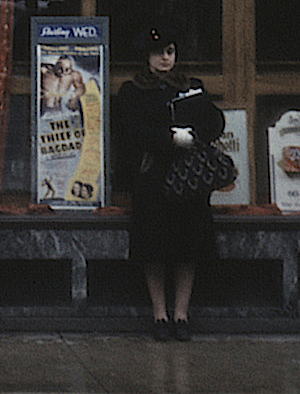}
            \caption{Polyblur~\cite{delbracio21polyblur}.}
        \end{subfigure} &
        \begin{subfigure}[t]{0.25\linewidth}
            \centering
            \includegraphics[width=\linewidth,trim=10 10 10 10,clip]{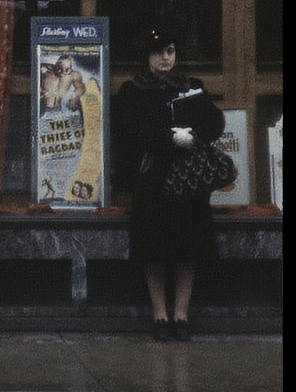}
            \caption{UNet-TX.}
        \end{subfigure} \\
    \end{tabular} 
    }
    \caption{Sharpening on the heritage image from \cite{schuler12blind}. The two classical methods have hyper-parameters to tune whereas our UNet-TX with $N=8$ has not and produce a halo-free sharp result.}
    \label{fig:sharpening}
\end{figure}

\paragraph{Sharpening.}
In the absence of any benchmark evaluating sharpening algorithms,
we simply run a qualitative comparison for sharpening.
Since mild blurs may be reasonably
approximated with Gaussians~\cite{eboli22breaking},
we select the lightweight UNet-TX trained in Section~\ref{subsec:ablation}
and compare it with Polyblur~\cite{delbracio21polyblur}
and the unsharp mask algorithm, two approaches
used on-device, for instance to postprocess multi-frame algorithms
like in the context of super-resolution or high-dynamic range imaging~\cite{lecouat22hdr}.
The comparison is run on a heritage photograph from \cite{schuler12blind}, and shown in Figure~\ref{fig:sharpening}.

\paragraph{Discussion and limitations.}
We have shown that the
hypothesis that similar blurs exist in blurry images and may be
gathered is verified, and leads to improvements in each situation.
Second, these improvements are obtained for a learning-free max pooling 
layer that leads to lightweight efficient models for sharpening and 
optical aberration correction. In these cases, $N=4$ or $N=8$ similar
patches are enough. For more diverse blur families such
as camera shake, learning-based pooling strategies and more important
stack sizes, {\em e.g.,} $N=16$ instead of 4, lead to significant
boost, suggesting that in this case, the more patches and pooling capacity, the better.

Nevertheless the proposed approach has limits.
Finding patches with similar blurs is not straightforward when the blur
may vary non-continuously, in 
particular with segmentation-aware blurs such as the dynamic blur  featured
in the GoPro dataset~\cite{nah17multiscale}, or depth-depending defocus blur~\cite{hasinoff07layerbased}. 
Another problem may arise if not enough patches with the same blur are collected. 
We remarked during our experiments that important noise alters the 
collaboration, thus limiting the benefits of our approach in high-noise regimes.
Yet, deblurring in such regime is an objective~\cite{anger19l0} 
beyond the scope of this paper. We have shown that for
three realistic blurring scenarios our method was effective.

\section{Conclusion}
\label{sec:conclusion}

In this paper, we have presented a simple modification of existing
CNNs for enhanced blind image deblurring. It consists in processing
together images with the same latent blur to share the different 
features of the blur across the images within   a network
which has internal collaborative layers taking the form
of feature maps pooling.
Finding several images having similar latent blurs is verified to be possible for a wide range of practical 
single-image blind deblurring applications. 
Experiments on
both synthetic and real-world images covering camera shake removal,
optical aberration compensation and sharpening validate our approach
and highlight the versatility of possible collaborating layers
to design efficient models.
This seems to establish  a practical framework upon which building both
lightweight and state-of-the-art architectures.

\paragraph{Aknowledgemnts.}
This work was partly financed by the DGA Astrid Maturation project SURECAVI ANR-21-ASM3-0002, the Office of Naval research grant N00014-17-1-2552, and the ANR project IMPROVED ANR-22-CE39-0006-04. This work was performed using HPC resources from GENCI–IDRIS (grants 2023-AD011012453R2, 2023-AD011012458R2). 
Centre Borelli is also with Université Paris Cité, SSA and INSERM.
The authors thank Adrien Courtois for fruitful discussions and help with the learnable pooling layers' implementations.

%%%%%%%%% REFERENCES
{\small
\bibliographystyle{plain}
\bibliography{bibliography}

\begin{thebibliography}{10}

\bibitem{abuolaim20dualpixel}
Abdullah Abuolaim and Michael~S. Brown.
\newblock Defocus deblurring using dual-pixel data.
\newblock In {\em European Conference on Computer Vision}, pages 111--126.
  Springer, 2020.

\bibitem{agustsson17div2k}
Eirikur Agustsson and Radu Timofte.
\newblock {NTIRE} 2017 challenge on single image super-resolution: Dataset and
  study.
\newblock In {\em {IEEE} Conference on Computer Vision and Pattern Recognition
  Workshops}, pages 1122--1131, 2017.

\bibitem{aittala18burst}
Miika Aittala and Fr{\'{e}}do Durand.
\newblock Burst image deblurring using permutation invariant convolutional
  neural networks.
\newblock In {\em European Conference on Computer Vision}, pages 748--764.
  Springer, 2018.

\bibitem{anger19l0}
J{\'{e}}r{\'{e}}my Anger, Gabriele Facciolo, and Mauricio Delbracio.
\newblock Blind image deblurring using the $\ell_0$ gradient prior.
\newblock {\em Image Processing On Line}, 9:124--142, 2019.

\bibitem{bauer18automatic}
Matthias Bauer, Valentin Volchkov, Michael Hirsch, and Bernhard
  Sch{\"{o}}lkopf.
\newblock Automatic estimation of modulation transfer functions.
\newblock In {\em {IEEE} International Conference on Computational
  Photography}, pages 1--12, 2018.

\bibitem{bello21lambda}
Irwan Bello.
\newblock Lambda{N}etworks: Modeling long-range interactions without attention.
\newblock In {\em International Conference on Learning Representations}, pages
  1--14. OpenReview.net, 2021.

\bibitem{brooks19unprocessing}
Tim Brooks, Ben Mildenhall, Tianfan Xue, Jiawen Chen, Dillon Sharlet, and
  Jonathan~T. Barron.
\newblock Unprocessing images for learned raw denoising.
\newblock In {\em {IEEE/CVF} Conference on Computer Vision and Pattern
  Recognition}, pages 11036--11045, 2019.

\bibitem{chen2008determining}
Mo~Chen, Jessica Fridrich, Miroslav Goljan, and Jan Luk{\'a}s.
\newblock Determining image origin and integrity using sensor noise.
\newblock {\em IEEE Transactions on information forensics and security},
  3(1):74--90, 2008.

\bibitem{chen21postprocessing}
Shiqi Chen, Huajun Feng, Dexin Pan, Zhihai Xu, Qi~Li, and Yue{-}ting Chen.
\newblock Optical aberrations correction in postprocessing using imaging
  simulation.
\newblock {\em {ACM} Transactions on Graphics}, 40(5):192:1--192:15, 2021.

\bibitem{cho21rethinking}
Sung{-}Jin Cho, Seo{-}Won Ji, Jun{-}Pyo Hong, Seung{-}Won Jung, and Sung{-}Jea
  Ko.
\newblock Rethinking coarse-to-fine approach in single image deblurring.
\newblock In {\em {IEEE/CVF} International Conference on Computer Vision},
  pages 4621--4630, 2021.

\bibitem{chu22improving}
Xiaojie Chu, Liangyu Chen, Chengpeng Chen, and Xin Lu.
\newblock Improving image restoration by revisiting global information
  aggregation.
\newblock In {\em European Conference on Computer Vision}, pages 53--71.
  Springer, 2022.

\bibitem{couzinie13learning}
Florent Couzinie{-}Devy, Jian Sun, Karteek Alahari, and Jean Ponce.
\newblock Learning to estimate and remove non-uniform image blur.
\newblock In {\em {IEEE} Conference on Computer Vision and Pattern
  Recognition}, pages 1075--1082, 2013.

\bibitem{dabov07bm3d}
Kostadin Dabov, Alessandro Foi, Vladimir Katkovnik, and Karen~O. Egiazarian.
\newblock Image denoising by sparse 3-{D} transform-domain collaborative
  filtering.
\newblock {\em {IEEE} Transactions on Image Processing}, 16(8):2080--2095,
  2007.

\bibitem{delbracio21polyblur}
Mauricio Delbracio, Ignacio Garcia{-}Dorado, Sungjoon Choi, Damien Kelly, and
  Peyman Milanfar.
\newblock Polyblur: Removing mild blur by polynomial reblurring.
\newblock {\em {IEEE} Transactions on Computational Imaging}, 7:837--848, 2021.

\bibitem{delbracio2023inversion}
Mauricio Delbracio and Peyman Milanfar.
\newblock Inversion by direct iteration: An alternative to denoising diffusion
  for image restoration.
\newblock {\em arXiv preprint arXiv:2303.11435}, 2023.

\bibitem{delbracio15fba}
Mauricio Delbracio and Guillermo Sapiro.
\newblock Removing camera shake via weighted {F}ourier burst accumulation.
\newblock {\em {IEEE} Transactions on Image Processing}, 24(11):3293--3307,
  2015.

\bibitem{eboli22breaking}
Thomas Eboli, Jean{-}Michel Morel, and Gabriele Facciolo.
\newblock Breaking down polyblur: Fast blind correction of small anisotropic
  blurs.
\newblock {\em Image Processing On Line}, 12:435--456, 2022.

\bibitem{eboli22fast}
Thomas Eboli, Jean-Michel Morel, and Gabriele Facciolo.
\newblock Fast two-step blind optical aberration correction.
\newblock In {\em European Conference on Computer Vision}, pages 693--708.
  Springer, 2022.

\bibitem{gao2019dynamic}
Hongyun Gao, Xin Tao, Xiaoyong Shen, and Jiaya Jia.
\newblock Dynamic scene deblurring with parameter selective sharing and nested
  skip connections.
\newblock In {\em {IEEE/CVF} Conference on Computer Vision and Pattern
  Recognition}, pages 3848--3856, 2019.

\bibitem{goldstein12blur}
Amit Goldstein and Raanan Fattal.
\newblock Blur-kernel estimation from spectral irregularities.
\newblock In {\em European Conference on Computer Vision}, pages 622--635.
  Springer, 2012.

\bibitem{hasinoff07layerbased}
Samuel~W. Hasinoff and Kiriakos~N. Kutulakos.
\newblock A layer-based restoration framework for variable-aperture
  photography.
\newblock In {\em {IEEE} International Conference on Computer Vision}, pages
  1--8, 2007.

\bibitem{zhe15regions}
Zhe Hu and Ming{-}Hsuan Yang.
\newblock Learning good regions to deblur images.
\newblock {\em International Journal on Computer Vision}, 115(3):345--362,
  2015.

\bibitem{kee11modeling}
Eric Kee, Sylvain Paris, Simon Chen, and Jue Wang.
\newblock Modeling and removing spatially-varying optical blur.
\newblock In {\em {IEEE} International Conference on Computational
  Photography}, pages 1--8, 2011.

\bibitem{kupyn19deblurgan}
Orest Kupyn, Tetiana Martyniuk, Junru Wu, and Zhangyang Wang.
\newblock Deblur{GAN}-v2: Deblurring (orders-of-magnitude) faster and better.
\newblock In {\em {IEEE/CVF} International Conference on Computer Vision},
  pages 8877--8886, 2019.

\bibitem{lafenetre23handheld}
Jamy Lafenetre, Gabriele Facciolo, and Thomas Eboli.
\newblock Implementing handheld burst super-resolution.
\newblock {\em Image Processing On Line}, 13, 2023.

\bibitem{lai22face}
Wei{-}Sheng Lai, Yichang Shih, Lun{-}Cheng Chu, Xiaotong Wu, Sung{-}Fang Tsai,
  Michael Krainin, Deqing Sun, and Chia{-}Kai Liang.
\newblock Face deblurring using dual camera fusion on mobile phones.
\newblock {\em {ACM} Transactions on Graphics}, 41(4):148:1--148:16, 2022.

\bibitem{lalonde2010sun}
Jean-Fran{\c{c}}ois Lalonde, Srinivasa~G Narasimhan, and Alexei~A Efros.
\newblock What do the sun and the sky tell us about the camera?
\newblock {\em International Journal of Computer Vision}, 88(1):24--51, 2010.

\bibitem{lecouat22hdr}
Bruno Lecouat, Thomas Eboli, Jean Ponce, and Julien Mairal.
\newblock High dynamic range and super-resolution from raw image bursts.
\newblock {\em {ACM} Transactions on Graphics}, 41(4):38:1--38:21, 2022.

\bibitem{levin09understanding}
Anat Levin, Yair Weiss, Fr{\'{e}}do Durand, and William~T. Freeman.
\newblock Understanding and evaluating blind deconvolution algorithms.
\newblock In {\em {IEEE} Conference on Computer Vision and Pattern
  Recognition}, pages 1964--1971, 2009.

\bibitem{liu21swin}
Ze~Liu, Yutong Lin, Yue Cao, Han Hu, Yixuan Wei, Zheng Zhang, Stephen Lin, and
  Baining Guo.
\newblock Swin transformer: Hierarchical vision transformer using shifted
  windows.
\newblock In {\em {IEEE/CVF} International Conference on Computer Vision},
  pages 9992--10002, 2021.

\bibitem{ma22searching}
Karima Ma, Micha{\"{e}}l Gharbi, Andrew Adams, Shoaib Kamil, Tzu{-}Mao Li,
  Connelly Barnes, and Jonathan Ragan{-}Kelley.
\newblock Searching for fast demosaicking algorithms.
\newblock {\em {ACM} Transactions on Graphics}, 41(5):172:1--172:18, 2022.

\bibitem{michaeli14patch}
Tomer Michaeli and Michal Irani.
\newblock Blind deblurring using internal patch recurrence.
\newblock In {\em European Conference on Computer Vision}, pages 783--798.
  Springer, 2014.

\bibitem{nah17multiscale}
Seungjun Nah, Tae~Hyun Kim, and Kyoung~Mu Lee.
\newblock Deep multi-scale convolutional neural network for dynamic scene
  deblurring.
\newblock In {\em {IEEE} Conference on Computer Vision and Pattern
  Recognition}, pages 257--265, 2017.

\bibitem{pan18dark}
Jinshan Pan, Deqing Sun, Hanspeter Pfister, and Ming{-}Hsuan Yang.
\newblock Deblurring images via dark channel prior.
\newblock {\em {IEEE} Transactions on Pattern Analysis and Machine
  Intelligence}, 40(10):2315--2328, 2018.

\bibitem{park20multitemporal}
Dongwon Park, Dong~Un Kang, Jisoo Kim, and Se~Young Chun.
\newblock Multi-temporal recurrent neural networks for progressive non-uniform
  single image deblurring with incremental temporal training.
\newblock In {\em European Conference on Computer Vision}, pages 327--343.
  Springer, 2020.

\bibitem{park14gyro}
Sung~Hee Park and Marc Levoy.
\newblock Gyro-based multi-image deconvolution for removing handshake blur.
\newblock In {\em {IEEE} Conference on Computer Vision and Pattern
  Recognition}, pages 3366--3373, 2014.

\bibitem{ranftl21transformer}
Ren{\'{e}} Ranftl, Alexey Bochkovskiy, and Vladlen Koltun.
\newblock Vision transformers for dense prediction.
\newblock In {\em {IEEE/CVF} International Conference on Computer Vision},
  pages 12159--12168, 2021.

\bibitem{rim20dataset}
Jaesung Rim, Haeyun Lee, Jucheol Won, and Sunghyun Cho.
\newblock Real-world blur dataset for learning and benchmarking deblurring
  algorithms.
\newblock In {\em European Conference on Computer Vision}, pages 184--201.
  Springer, 2020.

\bibitem{schuler12blind}
Christian~J. Schuler, Michael Hirsch, Stefan Harmeling, and Bernhard
  Sch{\"{o}}lkopf.
\newblock Blind correction of optical aberrations.
\newblock In {\em European Conference on Computer Vision}, pages 187--200.
  Springer, 2012.

\bibitem{tao18scale}
Xin Tao, Hongyun Gao, Xiaoyong Shen, Jue Wang, and Jiaya Jia.
\newblock Scale-recurrent network for deep image deblurring.
\newblock In {\em {IEEE} Conference on Computer Vision and Pattern
  Recognition}, pages 8174--8182, 2018.

\bibitem{vaswani2017attention}
Ashish Vaswani, Noam Shazeer, Niki Parmar, Jakob Uszkoreit, Llion Jones,
  Aidan~N. Gomez, Lukasz Kaiser, and Illia Polosukhin.
\newblock Attention is all you need.
\newblock In {\em Advances in Neural Information Processing Systems}, pages
  5998--6008, 2017.

\bibitem{wang22uformer}
Zhendong Wang, Xiaodong Cun, Jianmin Bao, Wengang Zhou, Jianzhuang Liu, and
  Houqiang Li.
\newblock Uformer: {A} general {U}-shaped transformer for image restoration.
\newblock In {\em {IEEE/CVF} Conference on Computer Vision and Pattern
  Recognition}, pages 17662--17672, 2022.

\bibitem{whang22refinement}
Jay Whang, Mauricio Delbracio, Hossein Talebi, Chitwan Saharia, Alexandros~G.
  Dimakis, and Peyman Milanfar.
\newblock Deblurring via stochastic refinement.
\newblock In {\em {IEEE/CVF} Conference on Computer Vision and Pattern
  Recognition}, pages 16272--16282, 2022.

\bibitem{whyte12shake}
Oliver Whyte, Josef Sivic, Andrew Zisserman, and Jean Ponce.
\newblock Non-uniform deblurring for shaken images.
\newblock {\em International Journal on Computer Vision}, 98(2):168--186, 2012.

\bibitem{li13unnatural}
Li~Xu, Shicheng Zheng, and Jiaya Jia.
\newblock Unnatural {$L_0$} sparse representation for natural image deblurring.
\newblock In {\em {IEEE} Conference on Computer Vision and Pattern
  Recognition}, pages 1107--1114, 2013.

\bibitem{zhang00calibration}
Zhengyou Zhang.
\newblock A flexible new technique for camera calibration.
\newblock {\em {IEEE} Transactions on Pattern Analysis and Machine
  Intelligence}, 22(11):1330--1334, 2000.

\end{thebibliography}
}

%% SUPMAT
\newpage

\appendix

\section{Broader impact}

The proposed technique aims at improving blind image deblurring. We are aware
that image processing may be used for non-ethical goals such as face recognition.
However, we believe that the technology is not mature enough to be used as is 
in a surveillance system for instance, and is thus not harmful. We however think
it can be useful for personal photography or certain scientific fields where
imagery plays a key role.

\section{Training details}

We provide in this section additional details on the 
training protocols of all the tasks and all the quantitative results for all experiments: PSNR, SSIM and LPIPS.

\subsection{Pooling layer $p$ implementations}

We detail in the main paper three approaches to implement the pooling function
$p$ in Eq.~\eqref{eq:pooling}: max pooling, lambda layer~\cite{bello21lambda}
and the self-attention module~\cite{vaswani2017attention}. We provide
formal definitions of these layers below.
We recall Equations~\eqref{eq:pooling} and \eqref{eq:conv1x1} in the main paper:
\begin{equation}
    g = p(\{e_1, \dots, e_N\}),
\end{equation}
and 
\begin{equation}
    e_n \gets \texttt{conv}_{1\times1}(\{e_n, g\}), \quad\forall n \in \{1,\dots,N\}.
\end{equation}
For the sake of explanation in this part, each feature map $e_n$
($n=1,\dots,N$) and the global feature map $g$ have dimension $H\times W\times C$.
We provide in what follows the python codes of 
our pooling modules. In all the codes we provide, the output array $g$ is of shape $(BN,C,H,W)$. The additional $N$
shape is for merging in a batch manner $g$ with the $N$
intermediate feature vectors within the subsequent $1\times1$
convolutions layers.
We first need in preamble of these codes
a few standard layers:

\begin{lstlisting}[language=Python, caption=Layer Norm and Layer Scale layers.]

def _pad_as(x, ref):
    _, _, *dims = ref.size()

    for _ in range(len(dims)):
        x = x.unsqueeze(dim=-1)

    return x


class LayerNorm(nn.Module):
    def __init__(self, in_channels: int, eps: float = 1e-6) -> None:
        super().__init__()

        self.eps = eps
        self.weight = nn.Parameter(torch.ones(in_channels))
        self.bias = nn.Parameter(torch.zeros(in_channels))

    def forward(self, x: torch.Tensor) -> torch.Tensor:
        u = x.mean(1, keepdim=True)
        s = (x - u).pow(2).mean(1, keepdim=True)
        x = (x - u) / torch.sqrt(s + self.eps)

        return torch.einsum("c, b c ... -> b c ...", self.weight, x) + _pad_as(self.bias, x)


class LayerScale(nn.Module):
    def __init__(self, in_channels: int, init: Optional[float] = 0.1):
        super().__init__()

        self.init = init

        if init is not None:
            self.weight = nn.Parameter(init * torch.ones(in_channels))

    def forward(self, x):
        if self.init is not None:
            return torch.einsum("b c ..., c -> b c ...", x, self.weight)

        return x
\end{lstlisting}

\paragraph{Max pooling.} As in \cite{aittala18burst}, we may simply implement $p$ as an
entrywise max operation.

\begin{lstlisting}[language=Python, caption=Max pooling module]
class PixelwiseMax(nn.Module):
    def __init__(self, N: int) -> None:
        super().__init__()

        self.N = N

    def forward(self, x: torch.Tensor) -> torch.Tensor:
        """
        Input: torch.Tensor of dimension [B * N, C, H, W]
        Output: torch.Tensor of dimension [B * N, C, H, W]
        """

        BN, C, H, W = x.size()  # (BN,C,H,W)
        B = BN // self.N

        # Pooling on the stack N axis.
        x_ = x.reshape(B, -1, C, H, W)
        g, _ = torch.max(x_, dim=1, keepdim=True)  # (B,1,C,H,W)
        g = g.repeat(1, self.N, 1, 1, 1).reshape(BN,C,H,W)

        return g
\end{lstlisting}

\paragraph{Self-attention module~\cite{vaswani2017attention}.}
We may implement the self-attention mechanism as defined in \cite{vaswani2017attention} to propose a learning-based
variant to the max pooling that finds interesting information
to extract from a sequence of feature maps.

\begin{lstlisting}[language=Python, caption=Self-attention pooling module]

class PixelwiseSAUp(nn.Module):
    def __init__(self, in_channels: int, N: int, heads: Optional[int] = None, 
                 layerscale_init: Optional[float] = 0.1) -> None:
        super().__init__()

        self.N = N
        self.heads = heads or in_channels // 32

        self.prenorm = LayerNorm(in_channels)
        self.to_QKV = nn.Conv2d(in_channels, 3 * in_channels, 1)
        self.layerscale = LayerScale(in_channels, init=layerscale_init)

        self.mlp = nn.Sequential(
            LayerNorm(in_channels),
            nn.Conv2d(in_channels, 4 * in_channels, 1),
            nn.GELU(),
            nn.Conv2d(4 * in_channels, in_channels, 1),
            LayerScale(in_channels, init=layerscale_init)
        )

    def forward(self, x: torch.Tensor) -> torch.Tensor:
        """
        Input: torch.Tensor of dimension [B * N, C, H, W]
        Output: torch.Tensor of dimension [B * N, C, H, W]
        """

        BN, C, H, W = x.size()
        B = BN // self.N

        # Query-Key-Value decomposition.
        QKV = self.to_QKV(self.prenorm(x))
        Q, K, V = torch.chunk(QKV, 3, dim=1)
        Q, K, V = map(lambda x: x.view(B, self.N, self.heads, -1, H, W), (Q, K, V))

        # Generate attention matrix.
        A = torch.einsum("b n h c i j, b m h c i j -> b n m h i j", Q, K) 
        A = A / sqrt(Q.size(3))
        A = A.softmax(2)

        # Apply the attention matrix.
        g = torch.einsum("b n m h i j, b m h c i j -> b n h c i j", A, V)
        g = g.reshape(BN, C, H, W)

        # Refining MLP.
        g = x + self.layerscale(g)
        g = g + self.mlp(g)
    
        return g
\end{lstlisting}

\paragraph{Lambda layer~\cite{bello21lambda}.}
the Lambda layer as proposed in \cite{bello21lambda}
is a sort of lightweight attention-like mechanism
that is a valid candidate for implementing
$p$. In the main paper for camera shake, we 
see in Table~\ref{tab:motion} that it leads to results better
than max pooling and lower than SA, but for a smaller
additional number of parameter.

\begin{lstlisting}[language=Python, caption=Lambda pooling module]
class PixelwiseLambdaUp(nn.Module):
    def __init__(self, in_channels: int, k: int, N: int, layerscale_init: Optional[float] = 0.1) -> None:
        super().__init__()

        self.N = N  # Number of frames in the stack.
        self.k = k  # Number of features in the Lambda layer.

        self.prenorm = LayerNorm(in_channels)
        self.to_K = nn.Conv2d(in_channels, k, 1)
        self.to_Q = nn.Conv2d(in_channels, in_channels, 1)
        self.merge_inner = nn.Conv2d((k + 1) * in_channels, in_channels, 1)
        self.layerscale = LayerScale(in_channels, init=layerscale_init)

    def forward(self, x: torch.Tensor) -> torch.Tensor:
        """
        Input: torch.Tensor of dimension [B * N, C, H, W]
        Output: torch.Tensor of dimension [B * N, C, H, W]
        """

        BN, _, H, W = x.size()
        B = BN // self.N

        # Query-Key decomposition on k features.
        x_norm = self.prenorm(x)
        Q = self.to_Q(x_norm)
        K = self.to_K(x_norm)
        Q, K = map(lambda x: x.view(B, self.N, -1, H, W), (Q, K))

        # Merge of the k features in a single array.
        K = K.softmax(1)
        output = torch.einsum('b n k i j, b n c i j -> b k c i j', K, Q)
        
        # Concatenate output to input
        output = output.reshape(B, 1, -1, H, W)
        output = output.repeat(1, self.N, 1, 1, 1)
        output = output.reshape(-1, output.size(2), H, W)
        
        g = torch.cat((x, output), 1)
        g = self.merge_inner(g)
        g = self.layerscale(g)

        return g
\end{lstlisting}

\subsection{Architectures of UNet and UNet-T}

\begin{figure}[h!]
    \centering
    \includegraphics[width=\linewidth]{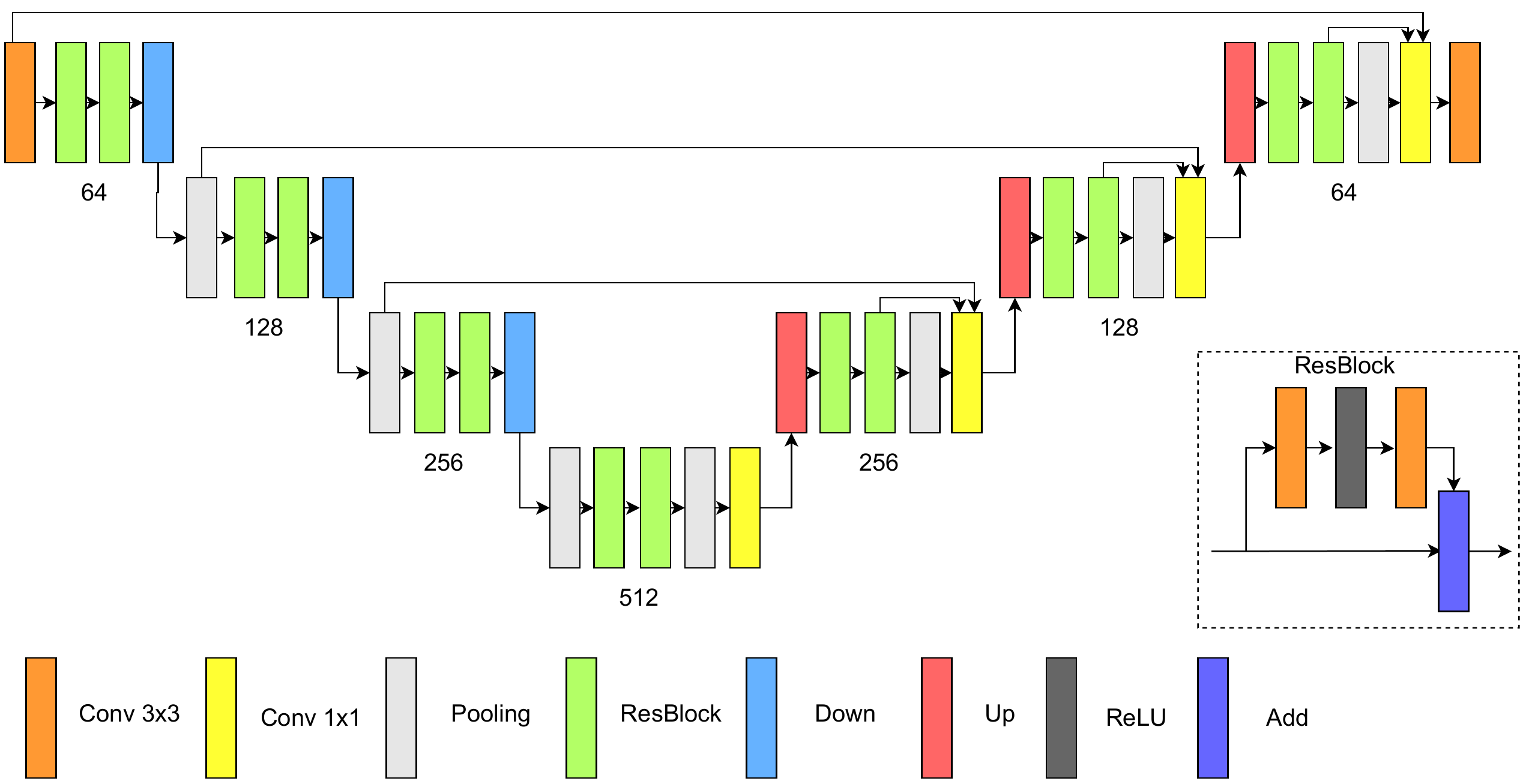}
    \caption{Diagram of UNet.}
    \label{fig:unet}
\end{figure}

\begin{figure}[h!]
    \centering
    \includegraphics[width=0.8\linewidth]{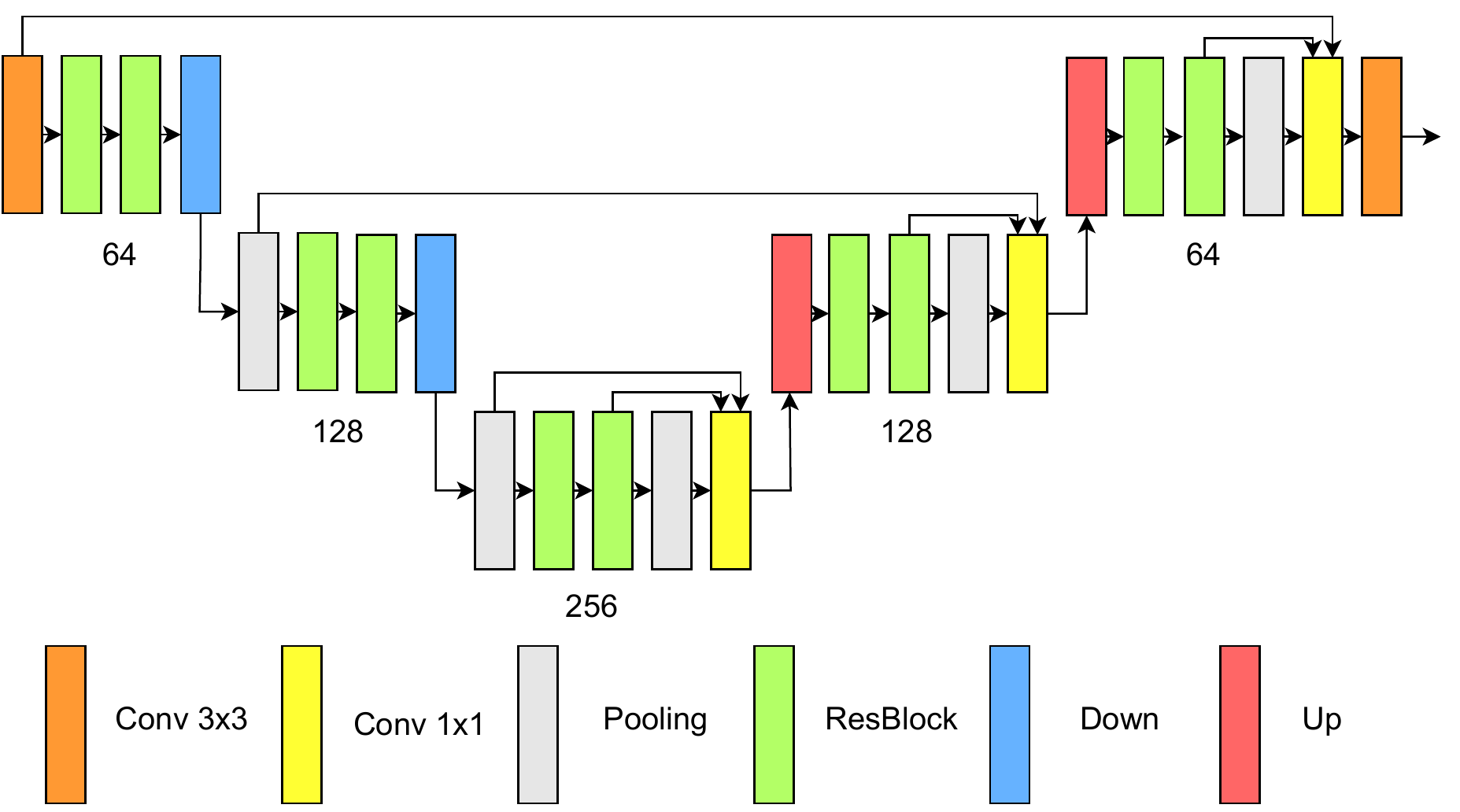}
    \caption{Diagram of UNet-T.}
    \label{fig:unetT}
\end{figure}

We provide diagrams in Figures~\ref{fig:unet}
and \ref{fig:unetT} of the considered 
UNet and UNet-T architectures respectively.
To get the architectures of UNet-S and UNet-X, divide
the channel widths shown in the Figure by 2 and 4 respectively. The same goes to get UNet-TS and UNet-TX
from UNet-T. We have not exhaustively searched
for an efficient architecture. Other configurations
embedding the pooling layer may lead to better or
worse results.
When $N=1$, the pooling module (gray box) in
Figures~\ref{fig:unet} and \ref{fig:unetT}
is removed, yielding a classical UNet merging
the contributions of the upsampled features and the 
ones in the encoder at the same scale.

\subsection{Slicing the images per type of blur}
\label{sec:slicing}

\paragraph{Camera shake.}
We focus on the Realblur-J~\cite{rim20dataset} images that are roughly
of size $700\times700$. When we restore an image of the test set 
(like in Table~\ref{tab:motion}), we simply slice the image into overlapping
patches with $25\%$ of overlap. Empirically, we have noticed that
since the images are downsampled versions of 12Mp photographs, each patch
has a sufficient different content from the other ones. In the draft, we
have limited ourselves to $N=1$, $4$ or $16$ but in practice even
more patches may be considered.

\paragraph{Sharpening.}
We follow the same strategy as for camera shake
since the blur that should be removed by a sharpening
stage such as fusion blur~\cite{eboli22breaking} or mild blur~\cite{delbracio21polyblur} are a sort of Gaussian
blur that is similar across the image field of view.
We thus first slice the image into overlapping patches
with overlap of $25\%$. We then feed stacks of $N$ patches
to the model, and when all the patches have been processed
by stack, we rearrange the patches into the full
image with the Bartlett window to eliminate the
fusion artifacts~\cite{schuler12blind}.

\paragraph{Optical aberration.}
Most images where optical aberration should be removed are either raw
images in 12, 24 or even 60Mpixel resolution~\cite{bauer18automatic, eboli22fast}, are large-enough JPEG images where the aberrations are
visible~\cite{kee11modeling, schuler12blind}.
Patch slicing is thus a viable strategy to handle the size of the images
and the non-uniformness of the blur~\cite{chen21postprocessing, eboli22fast}.

When provided an aberrated image, we select the patches
in the top-left quadrant of the image with respect of the
optical center, and select the three corresponding crops
with respect to the horizontal and vertical symmetries. 
Other symmetries could be exploited like diagonal symmetries
but we have not explored that direction.

\subsection{Training protocols}

For the sake of reproducibility, we present here all the the training details
that may have not fit into the main paper.

\paragraph{Gaussian blur.}
We select the 800 training images of the DIV2K dataset~\cite{agustsson17div2k}. We randomly crop $128\times128$ RGB patches from these images,
unprocess them with the protocol of \cite{brooks19unprocessing} (inverse tone mapping, gamma expansion, and inverse color matrix), blur the images with 
Gaussian blurs with standard deviations along the two
principal axes in $[0.3,4]^2$ and orientation randomly
sampled in $[0,2\pi]$, add Gaussian noise with standard
deviation in $[0.5,2]/255$. We then reprocess the image 
(color matrix, gamma compression, and tone mapping). This mimics that RGB images are generally shown in a domain where the blur kernel has not a simple linear relationship
with the blurry image at end because of a camera ISP.

For each training stack of size $N$, we randomly select a single
Gaussian blur and noise level, and an
image among the 800 at hand from which we crop $N$ patches in it,
and degrade them with the protocol shown above to generate the
input stack of $N$ blurry images sharing the same blur.

\paragraph{Optical aberration.}
We follow the same strategy as for Gaussian blur, but do not
reapply the camera color matrix, gamma compression and tone mapping
to be in a case where the correction is applied on a linear RGB
image since optical aberration correction is one
of the core tasks of an ISP pipeline~\cite{eboli22fast}.

Because of how are sampled the patches in a real aberrated images
(cf Section~\ref{sec:slicing}), we restrict the number of images $N$
in the stack to be either 1 or 4.
For $N=1$ and the given PSF chosen among the 70 from \cite{bauer18automatic}, we uniformly sample a location on the field
of view of the camera and select the nearest local $41\times41$ RGB filter.
It is applied to the $128\times128$ RGB random crop.
For $N=4$, we instead sample a location in the top-left quadrant of the field
of view of the camera, and fetch the 4 local RGB  filters 
following the vertical and horizontal symmetry. These local RGB
filters are all {\em similar} but not exactly the same, thus
training our model to handle the differences of support.

\paragraph{Camera shake.}
For camera shake removal in the Realblur-J dataset~\cite{rim20dataset},
we select an image, uniformly sample a $256\times256$ crop
and slice into into $N=4$ or $N=16$ {\em non-overlapping}
patches of the same size, {\em i.e.,} 4 patches of size $128\times128$ or 16 patches of size $64\times64$. The rationale behind this choice
is that we simply want to learn deblurring, not deblurring
an image like at test time. We found that this simple and fast
strategy was working well in practice since the estimated blur
kernels provided with the dataset are almost all of support smaller than $54\times64$ pixels, thus for all values of $N$, the patches
capture the whole support of the blur.

\section{Additional quantitative results}

We provide in this section the results for the three metrics PSNR, SSIM and LPIPS to complement the Tables~\ref{tab:ablationdepth} and \ref{tab:aberrations}, 
and Figure~\ref{fig:ablationwidth} where we present
results only for the average PSNR because of lack
of space.

\subsection{Gaussian blur (depth)}

The following tables complement Table~\ref{tab:ablationdepth} in the main paper where we reported only the average PSNR over 3 splits of the 
dataset. We show in Tables~\ref{tab:ablation_supmat_depth_psnr}, \ref{tab:ablation_supmat_depth_ssim} and \ref{tab:ablation_supmat_depth_lpips} the average PSNR, SSIM and LPIPS
scores for the 100 DIV2K test images used to evaluate the different models, per stack size $N$ and standard deviation of the isotropic
Gaussian blur $\sigma$. The average PSNR scores are those
shown in Table~\ref{tab:ablationdepth} in the main paper.
On the three metrics one may observe improvements with $N$, and saturation
happening around $N=8$, as noted in the main paper.

\begin{table}[h!]
    \centering
  \begin{tabular}{lcccccc}
    \toprule
     & $\sigma$ & 1 & 2 & 3 & 4 & Average\\
    \midrule
    UNet-T & $N=1$  & 35.59 & 32.67 & 31.25 & 30.35 & 32.46 \\
    UNet-T & $N=2$  & 36.05 & 33.04 & 31.44 & 30.46 & 32.75 \\
    UNet-T & $N=4$  & 36.31 & 33.20 & 31.56 & 30.58 & 32.92 \\
    UNet-T & $N=8$  & 36.70 & 33.44 & 31.73 & 30.66 & 33.13 \\
    UNet-T & $N=16$ & 36.59 & 33.34 & 31.59 & 30.45 & 33.02 \\
    \midrule
    UNet & $N=1$  & 35.97 & 33.01 & 31.41 & 30.61 & 32.75 \\
    UNet & $N=2$  & 36.43 & 33.33 & 31.55 & 30.64 & 32.99 \\
    UNet & $N=4$  & 36.51 & 33.46 & 31.64 & 30.70 & 33.08 \\
    UNet & $N=8$  & 36.62 & 33.51 & 31.76 & 30.71 & 33.15 \\
    UNet & $N=16$ & 36.51 & 33.46 & 31.53 & 30.48 & 33.07 \\
    \bottomrule
  \end{tabular}
  \captionof{table}{PSNR scores for UNet and UNet-T per standard deviation level $\sigma$, and with $p$ implemented with max pooling.}
  \label{tab:ablation_supmat_depth_psnr}
\end{table}

\begin{table}[h!]
    \centering
  \begin{tabular}{lcccccc}
    \toprule
     & $\sigma$ & 1 & 2 & 3 & 4 & Average\\
    \midrule
    UNet-T & $N=1$  & 0.960 & 0.923 & 0.896 & 0.859 & 0.910 \\
    UNet-T & $N=2$  & 0.961 & 0.926 & 0.898 & 0.873 & 0.915 \\
    UNet-T & $N=4$  & 0.963 & 0.928 & 0.899 & 0.876 & 0.917 \\
    UNet-T & $N=8$  & 0.965 & 0.930 & 0.902 & 0.878 & 0.919\\
    UNet-T & $N=16$ & 0.964 & 0.929 & 0.900 & 0.875 & 0.918 \\
    \midrule
    UNet & $N=1$  & 0.962 & 0.926 & 0.898 & 0.876 & 0.916 \\
    UNet & $N=2$  & 0.963 & 0.928 & 0.899 & 0.877 & 0.917 \\
    UNet & $N=4$  & 0.964 & 0.930 & 0.900 & 0.878 & 0.918 \\
    UNet & $N=8$  & 0.964 & 0.930 & 0.902 & 0.879 & 0.919 \\
    UNet & $N=16$ & 0.964 & 0.928 & 0.900 & 0.878 & 0.918 \\
    \bottomrule
  \end{tabular}
  \captionof{table}{SSIM scores for UNet and UNet-T per standard deviation level $\sigma$, and with $p$ implemented with max pooling.}
  \label{tab:ablation_supmat_depth_ssim}
\end{table}

\begin{table}[h!]
    \centering
  \begin{tabular}{lcccccc}
    \toprule
     & $\sigma$ & 1 & 2 & 3 & 4 & Average\\
    \midrule
    UNet-T & $N=1$  & 0.029 & 0.075 & 0.120 & 0.159 & 0.096 \\
    UNet-T & $N=2$  & 0.026 & 0.069 & 0.115 & 0.152 & 0.091 \\
    UNet-T & $N=4$  & 0.021 & 0.064 & 0.112 & 0.146 & 0.086 \\
    UNet-T & $N=8$  & 0.015 & 0.058 & 0.107 & 0.140 & 0.080 \\
    UNet-T & $N=16$ & 0.016 & 0.059 & 0.108 & 0.142 & 0.081 \\
    \midrule
    UNet & $N=1$  & 0.025 & 0.068 & 0.113 & 0.145 & 0.088 \\
    UNet & $N=2$  & 0.022 & 0.063 & 0.110 & 0.142 & 0.084 \\
    UNet & $N=4$  & 0.021 & 0.060 & 0.108 & 0.136 & 0.081 \\
    UNet & $N=8$  & 0.017 & 0.058 & 0.105 & 0.135 & 0.079 \\
    UNet & $N=16$ & 0.016 & 0.060 & 0.109 & 0.141 & 0.081 \\
    \bottomrule
  \end{tabular}
  \captionof{table}{LPIPS scores for UNet and UNet-T per standard deviation level $\sigma$, and with $p$ implemented with max pooling.}
  \label{tab:ablation_supmat_depth_lpips}
\end{table}

\subsection{Gaussian blur (width)}

The following tables complement Figure~\ref{fig:ablationwidth} in the main paper where we reported only the average PSNR over 3 splits of the 
dataset.
We show in Tables~\ref{tab:ablation_supmat_depth_psnr}, \ref{tab:ablation_supmat_depth_ssim} and \ref{tab:ablation_supmat_depth_lpips} the average PSNR, SSIM and LPIPS
scores for the 100 DIV2K test images used to evaluate the different models, per stack size $N$ and standard deviation of the isotropic
Gaussian blur $\sigma$. The average PSNR scores are those
shown in Figure~\ref{fig:ablationwidth} in the main paper.
For each width, the variant with $N=8$ achieves better results than the counterpart with
$N=1$, especially $\sigma$ set to 1 and 2, typical values for image sharpening~\cite{eboli22breaking}.

Notably, the performance of UNet-TX ($N=8$) is
similar to that of UNet-TS ($N=1$), and the
performance of UNet-TS ($N=8$) is superior to
that of UNet ($N=1$), despite the first models
being four times smaller than the second ones.
This is true for all three metrics, and thus
validates our claim that collaborative filtering
helps miniaturizing deblurring networks.

\begin{table}[h!]
    \centering
    \begin{tabular}{lcccccc}
        \toprule
        & $\sigma$ & 1 & 2 & 3 & 4 & Average \\
        \midrule
        UNet-TX & $N=1$ & 33.84 & 31.80 & 30.70 & 29.43 & 31.44\\ 
        UNet-TX & $N=8$ & 34.63 & 32.25 & 30.81 & 29.80 & 31.87\\ 
        UNet-TS & $N=1$ & 34.86 & 32.32 & 31.10 & 29.88 & 32.04 \\ 
        UNet-TS & $N=8$ & 35.80 & 33.03 & 31.40 & 30.36 & 32.65 \\ 
        UNet-T & $N=1$ & 35.59 & 32.67 & 31.25 & 30.35 & 32.47\\ 
        UNet-T & $N=8$ & 36.70 & 33.43 & 31.73 & 30.66 & 33.13\\ 
        \midrule
        UNet-X & $N=1$ & 34.22 & 32.04 & 30.83 & 29.79 & 31.72\\ 
        UNet-X & $N=8$ & 35.01 & 32.53 & 31.03 & 30.03 & 32.15\\ 
        UNet-S & $N=1$ & 35.34 & 32.65 & 31.21 & 30.30 & 32.38\\ 
        UNet-S & $N=8$ & 35.99 & 33.21 & 31.55 & 30.47 & 32.81\\ 
        UNet & $N=1$ & 35.97 & 33.01 & 31.41 & 30.61 & 32.75\\ 
        UNet & $N=8$ & 36.62 & 33.51 & 31.76 & 30.71 & 33.15 \\
        \bottomrule
    \end{tabular}
    \caption{Average PSNR to benchmark width of the models.}
    \label{tab:ablation_supmat_width_psnr}
\end{table}

\begin{table}[h!]
    \centering
    \begin{tabular}{lcccccc}
        \toprule
        & $\sigma$ & 1 & 2 & 3 & 4 & Average \\
        \midrule
        UNet-TX & $N=1$ & 0.946 & 0.912 & 0.886 & 0.851 & 0.899\\ 
        UNet-TX & $N=8$ & 0.951 & 0.917 & 0.889 & 0.862 & 0.905\\ 
        UNet-TS & $N=1$ & 0.954 & 0.919 & 0.892 & 0.860 & 0.906\\ 
        UNet-TS & $N=8$ & 0.960 & 0.926 & 0.897 & 0.873 & 0.914\\ 
        UNet-T & $N=1$ & 0.960 & 0.923 & 0.896 & 0.869 & 0.912\\ 
        UNet-T & $N=8$ & 0.965 & 0.930 & 0.902 & 0.878 & 0.919\\ 
        \midrule
        UNet-X & $N=1$ & 0.948 & 0.915 & 0.889 & 0.861 & 0.903\\ 
        UNet-X & $N=8$ & 0.954 & 0.920 & 0.891 & 0.867 & 0.908\\ 
        UNet-S & $N=1$ & 0.958 & 0.923 & 0.895 & 0.870 & 0.912\\ 
        UNet-S & $N=8$ & 0.960 & 0.927 & 0.899 & 0.876 & 0.916\\ 
        UNet & $N=1$ & 0.962 & 0.926 & 0.898 & 0.876 & 0.916\\ 
        UNet & $N=8$ & 0.964 & 0.930 & 0.902 & 0.879 & 0.919 \\
        \bottomrule
    \end{tabular}
    \caption{Average SSIM to benchmark width of the models.}
    \label{tab:ablation_supmat_width_ssim}
\end{table}

\begin{table}[h!]
    \centering
    \begin{tabular}{lcccccc}
        \toprule
         & $\sigma$ & 1 & 2 & 3 & 4 & Average \\
        \midrule
        UNet-TX & $N=1$ & 0.043 & 0.093 & 0.140 & 0.191 & 0.117\\ 
        UNet-TX & $N=8$ & 0.031 & 0.077 & 0.123 & 0.162 & 0.098\\ 
        UNet-TS & $N=1$ & 0.037 & 0.084 & 0.129 & 0.178 & 0.107\\ 
        UNet-TS & $N=8$ & 0.020 & 0.063 & 0.113 & 0.149 & 0.086\\ 
        UNet-T & $N=1$ & 0.029 & 0.075 & 0.120 & 0.159 & 0.096\\ 
        UNet-T & $N=8$ & 0.015 & 0.058 & 0.107 & 0.140 & 0.080\\ 
        \midrule
        UNet-X & $N=1$ & 0.038 & 0.082 & 0.126 & 0.172 & 0.105 \\ 
        UNet-X & $N=8$ & 0.026 & 0.069 & 0.116 & 0.151 & 0.091 \\ 
        UNet-S & $N=1$ & 0.028 & 0.074 & 0.119 & 0.156 & 0.094\\ 
        UNet-S & $N=8$ & 0.019 & 0.061 & 0.109 & 0.143 & 0.083\\ 
        UNet & $N=1$ & 0.025 & 0.068 & 0.113 & 0.145 & 0.088\\ 
        UNet & $N=8$ & 0.017 & 0.058 & 0.105 & 0.138 & 0.079 \\
        \bottomrule
    \end{tabular}
    \caption{Average LPIPS to benchmark width of the models.}
    \label{tab:ablation_supmat_width_lpips}
\end{table}

\subsection{Optical aberration}

We report the PSNR, SSIM and LPIPS for the two lenses benchmarked
in Table~\ref{tab:aberrations} in the main paper.
The results for the 16mm lens are in Table~\ref{tab:aberrations_supmat_16mm} and those of the 24mm
lens are reported in Table~\ref{tab:aberrations_supmat_24mm}.
We run the models on 3 different sets of patches to compute error bars
on the variety of sets of patches that may be collected in practice.
However in the main paper we only run the methods on a single arrangement of
the image patches (explaining the slight differences of average PSNRs).
The differences between $N=1$ and $N=4$ in these tables
and Table~\ref{tab:aberrations} in the main paper are 
approximately of the same order.
The error bar reported in Tables~\ref{tab:aberrations_supmat_16mm}
and \ref{tab:aberrations_supmat_24mm} are the standard deviation
of the metrics on the three runs.

We remark that the collaborative strategy is particularly
efficient for the 16mm lens that is of less quality than the
24mm one. This suggests that for lower-quality lenses, collaboration
is further helpful.

\begin{table}[h!]
    \centering
    \begin{tabular}{llccc}
        \toprule
        Location & Method & PSNR & SSIM & LPIPS \\
        \midrule
        Corner & UNet-TX ($N=1$) & $30.78\pm0.12$ & $0.814\pm0.003$ & $0.223\pm0.003$ \\
        Corner & UNet-TX ($N=4$, M) & $31.21\pm0.11$ & $0.826\pm0.003$ & $0.214\pm0.003$\\
        Corner & UNet-TS ($N=1$) & $31.76\pm0.12$ & $0.840\pm0.003$ & $0.196\pm0.003$ \\
        Corner & UNet-TS ($N=4$, M) & $32.01\pm0.11$ & $0.846\pm0.003$ & $0.186\pm0.003$ \\
        Corner & UNet-T ($N=1$) &  $32.66\pm0.12$ & $0.861\pm0.002$ & $\mathbf{0.166}\pm0.003$ \\
        Corner & UNet-T ($N=4$, M) & $\mathbf{32.86}\pm0.12$ & $\mathbf{0.864}\pm0.002$ & $\mathbf{0.164}\pm0.003$ \\
        \midrule
        Intermediate & UNet-TX ($N=1$) & $34.54\pm0.08$ & $0.906\pm0.001$ & $0.108\pm0.002$ \\
        Intermediate & UNet-TX ($N=4$, M) & $34.92\pm0.11$ & $0.911\pm0.001$ & $0.105\pm0.002$ \\
        Intermediate & UNet-TS ($N=1$) & $35.29\pm0.01$ & $0.917\pm0.001$ & $0.092\pm0.002$ \\
        Intermediate & UNet-TS ($N=4$, M) & $35.66\pm0.11$ & $0.921\pm0.001$ & $0.088\pm0.002$ \\
        Intermediate & UNet-T ($N=1$) & $36.03\pm0.10$ & $0.927\pm0.001$ & $0.078\pm0.002$ \\
        Intermediate & UNet-T ($N=4$, M) & $\mathbf{36.32}\pm0.10$ & $\mathbf{0.929}\pm0.001$ & $\mathbf{0.076}\pm0.002$ \\
        \midrule
        Center & UNet-TX ($N=1$) & $37.13\pm0.11$ & $0.954\pm0.001$ & $0.037\pm0.001$ \\
        Center & UNet-TX ($N=4$, M) & $37.72\pm0.10$ & $0.956\pm0.001$ & $0.035\pm0.001$ \\
        Center & UNet-TS ($N=1$) &  $38.22\pm0.11$ & $0.961\pm0.001$ & $0.028\pm0.001$ \\
        Center & UNet-TS ($N=4$, M) & $38.51\pm0.09$ & $0.961\pm0.001$ & $0.027\pm0.001$ \\
        Center & UNet-T ($N=1$) & $39.24\pm0.11$ & $\mathbf{0.966}\pm0.001$ & $\mathbf{0.021}\pm0.001$ \\
        Center & UNet-T ($N=4$, M) & $\mathbf{39.37}\pm0.09$ & $\mathbf{0.966}\pm0.001$ & $\mathbf{0.020}\pm0.001$ \\
        \bottomrule
    \end{tabular}
    \caption{Quantitative results for the Canon 16-35mm f/2.8 EI USM lens opened at f/2.8 and at 16mm~\cite{bauer18automatic}.}
    \label{tab:aberrations_supmat_16mm}
\end{table}

\begin{table}[h!]
    \centering
    \begin{tabular}{llllc}
        \toprule
        Location & Method & PSNR & SSIM & LPIPS \\
        \midrule
        Corner & UNet-TX ($N=1$) & $33.13\pm0.12$ & $0.881\pm0.002$ & $0.132\pm0.002$ \\
        Corner & UNet-TX ($N=4$, M) & $33.61 \pm 0.13$ & $0.893\pm0.002$ & $0.120\pm0.002$ \\
        Corner & UNet-TS ($N=1$) & $34.26\pm0.12$ & $0.903\pm0.002$ & $0.101\pm0.002$ \\
        Corner & UNet-TS ($N=4$, M) & $34.77\pm0.13$ & $0.911\pm0.002$ & $0.092\pm0.002$ \\
        Corner & UNet-T ($N=1$) & $35.05\pm0.13$ & $0.916\pm0.002$ & $0.085\pm0.002$ \\
        Corner & UNet-T ($N=4$, M) & $\mathbf{35.47}\pm0.12$ &  $\mathbf{0.921}\pm0.001$ & $\mathbf{0.082}\pm0.002$\\
        \midrule
        Intermediate & UNet-TX ($N=1$) & $35.08\pm0.13$ & $0.909\pm0.002$ & $0.107\pm0.002$\\
        Intermediate & UNet-TX ($N=4$, M) & $35.22\pm0.14$ & $0.911\pm0.002$ & $0.105\pm0.002$ \\
        Intermediate & UNet-TS ($N=1$) & $35.71\pm0.12$ & $0.918\pm0.001$ & $0.094\pm0.002$ \\
        Intermediate & UNet-TS ($N=4$, M) & $35.90\pm0.12$ & $0.920\pm0.001$ & $0.091\pm0.002$ \\
        Intermediate & UNet-T ($N=1$) & $\mathbf{36.38}\pm0.12$ & $\mathbf{0.927}\pm0.001$ & $\mathbf{0.081}\pm0.002$ \\
        Intermediate & UNet-T ($N=4$, M) & $\mathbf{36.39}\pm0.12$ & $\mathbf{0.926}\pm0.001$ & $\mathbf{0.081}\pm0.002$ \\
        \midrule
        Center & UNet-TX ($N=1$) & $36.82\pm0.09$ & $0.946\pm0.001$ & $0.051\pm0.001$\\
        Center & UNet-TX ($N=4$, M) & $37.04\pm0.07$ & $0.947\pm0.001$ & $0.050\pm0.001$\\
        Center & UNet-TS ($N=1$) & $38.06\pm0.09$ & $0.955\pm0.001$ & $0.037\pm0.001$ \\
        Center & UNet-TS ($N=4$, M) & $38.16\pm0.08$ & $0.956\pm0.001$ & $0.035\pm0.001$ \\
        Center & UNet-T ($N=1$) & $39.11\pm0.09$ & $\mathbf{0.962}\pm0.001$ & $\mathbf{0.025}\pm0.001$ \\
        Center & UNet-T ($N=4$, M) & $\mathbf{39.04}\pm0.08$ & $\mathbf{0.962}\pm0.000$ & $\mathbf{0.025}\pm0.001$\\
        \bottomrule
    \end{tabular}
    \caption{Quantitative results for the Canon 24mm f/1.4L USM lens opened at f/1.4~\cite{bauer18automatic}.}
    \label{tab:aberrations_supmat_24mm}
\end{table}

\section{Additional qualitative results}

\subsection{Sharpening}

\begin{figure}[h!]
    \centering
    \resizebox{\linewidth}{!}{
    \begin{tabular}{cccc}
        \begin{subfigure}[t]{0.25\linewidth}
            \centering
            \includegraphics[width=\linewidth,trim=90 20 40 0, clip]{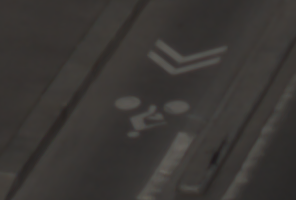}
            \caption{Blurry.}
        \end{subfigure} &
        \begin{subfigure}[t]{0.25\linewidth}
            \centering
            \includegraphics[width=\linewidth,trim=90 20 40 0, clip]{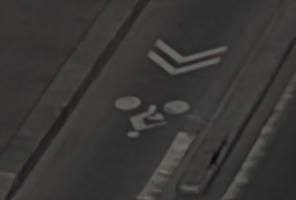}
            \caption{Unsharp mask.}
        \end{subfigure} &
        \begin{subfigure}[t]{0.25\linewidth}
            \centering
            \includegraphics[width=\linewidth,trim=90 20 40 0, clip]{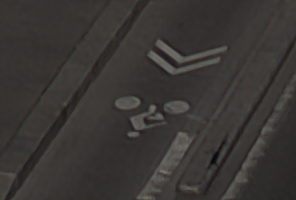}
            \caption{Polyblur~\cite{delbracio21polyblur}.}
        \end{subfigure} &
        \begin{subfigure}[t]{0.25\linewidth}
            \centering
            \includegraphics[width=\linewidth,trim=90 20 40 0, clip]{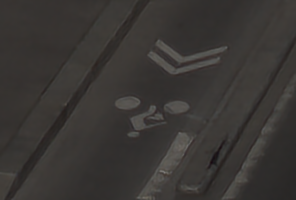}
            \caption{UNet-TX ($N=8$).}
        \end{subfigure} \\
    \end{tabular} 
    }
    \caption{Sharpening example taken from the ``Friant'' image obtained with the online demo of \cite{lafenetre23handheld}.}
    \label{fig:sharpening_cyclist}
\end{figure}

\begin{figure}[h!]
    \centering
    \resizebox{\linewidth}{!}{
    \begin{tabular}{cccc}
        \begin{subfigure}[t]{0.25\linewidth}
            \centering
            \includegraphics[width=\linewidth,trim=30 20 40 20, clip]{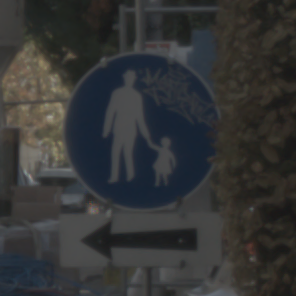}
            \caption{Blurry.}
        \end{subfigure} &
        \begin{subfigure}[t]{0.25\linewidth}
            \centering
            \includegraphics[width=\linewidth,trim=30 20 40 20, clip]{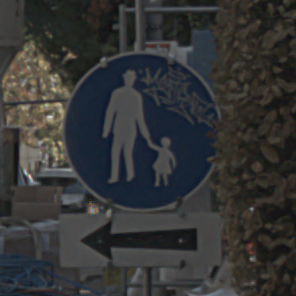}
            \caption{Unsharp mask.}
        \end{subfigure} &
        \begin{subfigure}[t]{0.25\linewidth}
            \centering
            \includegraphics[width=\linewidth,trim=30 20 40 20, clip]{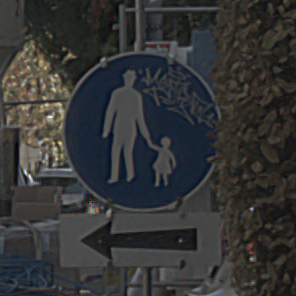}
            \caption{Polyblur~\cite{delbracio21polyblur}.}
        \end{subfigure} &
        \begin{subfigure}[t]{0.25\linewidth}
            \centering
            \includegraphics[width=\linewidth,trim=30 20 40 20, clip]{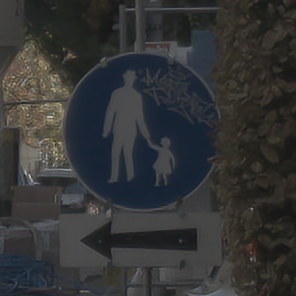}
            \caption{UNet-TX ($N=8$).}
        \end{subfigure} \\
    \end{tabular} 
    }
    \caption{Sharpening example taken from the ``Zurich'' image obtained with the online demo of \cite{lafenetre23handheld}.}
    \label{fig:sharpening_roadsign}
\end{figure}

We provide an additional sharpening example from the two
by-default images of the online super-resolution demo 
from~\cite{lafenetre23handheld}. 
We compare to the unsharp mask technique (classical
technique for sharpening), and Polyblur~\cite{delbracio21polyblur} (state-of-the-art for learning-free image sharpening). Since the official
implementation is not available, we use the non-official
one powering the online demo~\cite{eboli22breaking}.
The examples are shown in Figures~\ref{fig:sharpening_cyclist} and
\ref{fig:sharpening_roadsign}.
In general, our model does not suffer from halo artifacts
like unsharp mask, and has no hyper-parameter to adjust
like Polyblur to work well.
Our images may lack a bit of contrast enhancement
because we target Gaussian deblurring, a slightly
different problem to sharpening that also
seeks to improve the general contrast of the image.

\subsection{Camera shake}

\begin{figure}[h!]
    \centering
    \resizebox{\linewidth}{!}{
    \begin{tabular}{ccccc}
        \begin{subfigure}[t]{0.2\linewidth}
            \centering
            \includegraphics[width=\linewidth,trim=250 200 240 290, clip]{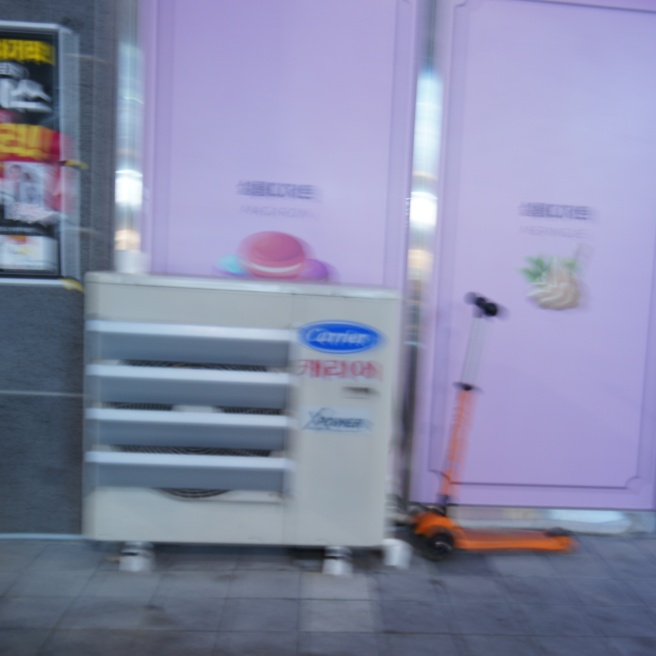}
            \caption{Blurry.}
        \end{subfigure} &
        \begin{subfigure}[t]{0.2\linewidth}
            \centering
            \includegraphics[width=\linewidth,trim=250 200 240 290, clip]{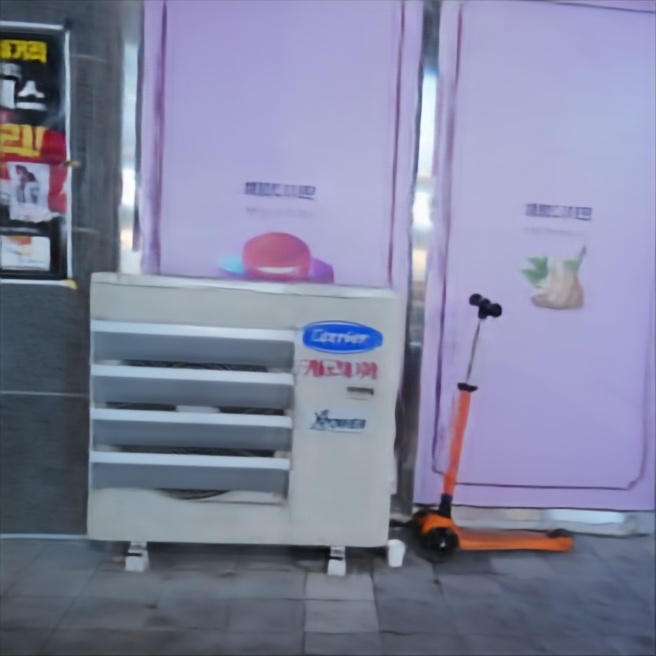}
            \caption{UNet-T.}
        \end{subfigure} &
        \begin{subfigure}[t]{0.2\linewidth}
            \centering
            \includegraphics[width=\linewidth,trim=250 200 240 290, clip]{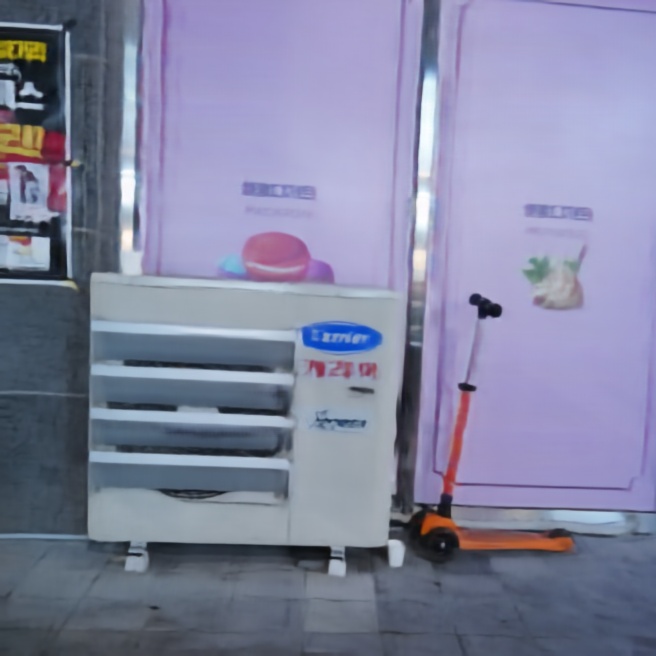}
            \caption{UNet-T (M, $4$).}
        \end{subfigure} &
        \begin{subfigure}[t]{0.2\linewidth}
            \centering
            \includegraphics[width=\linewidth,trim=250 200 240 290, clip]{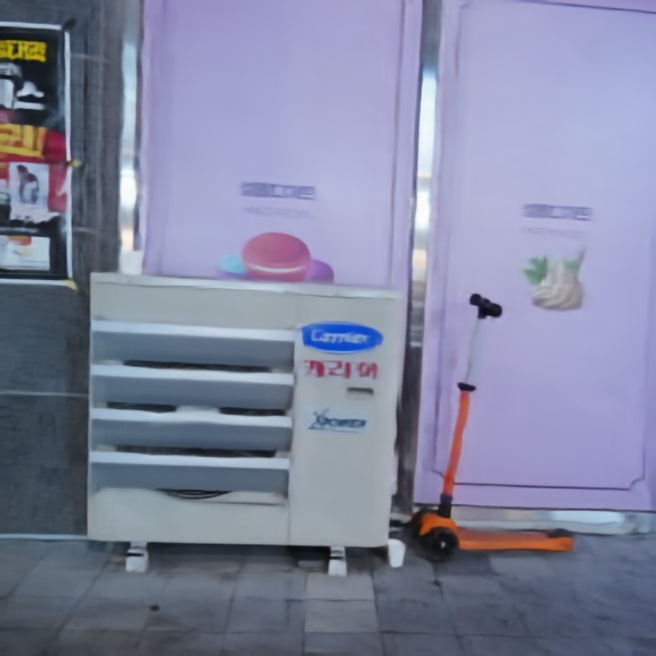}
            \caption{UNet-T (L, $4$).}
        \end{subfigure} &
        \begin{subfigure}[t]{0.2\linewidth}
            \centering
            \includegraphics[width=\linewidth,trim=250 200 240 290, clip]{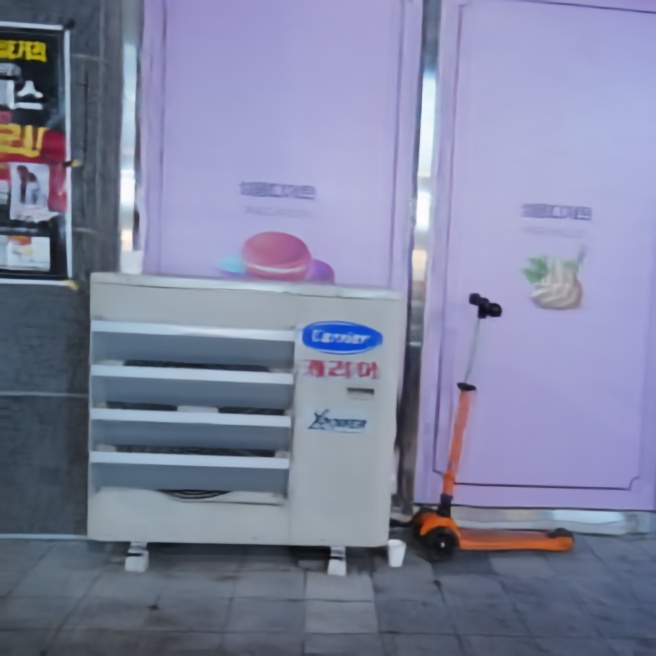}
            \caption{UNet-T (A, $4$).}
        \end{subfigure} \\
        \begin{subfigure}[t]{0.2\linewidth}
            \centering
            \includegraphics[width=\linewidth,trim=250 200 240 290, clip]{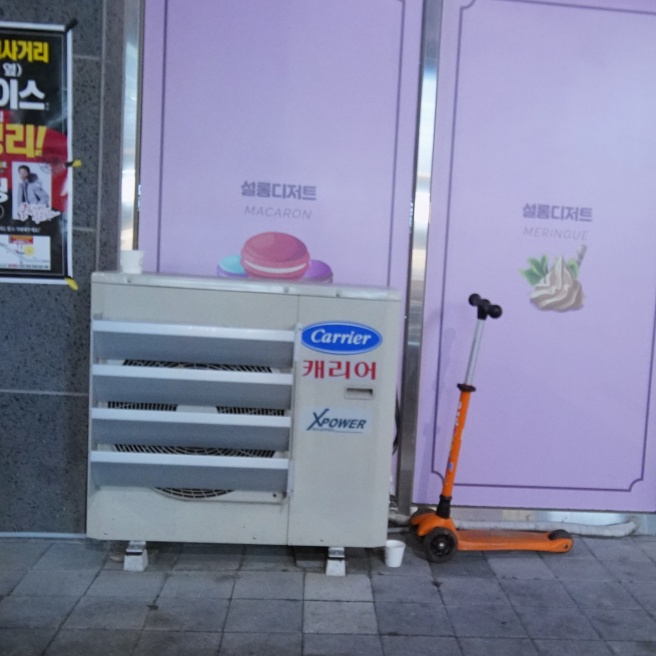}
            \caption{Target.}
        \end{subfigure} &
        \begin{subfigure}[t]{0.2\linewidth}
            \centering
            \includegraphics[width=\linewidth,trim=250 200 240 290, clip]{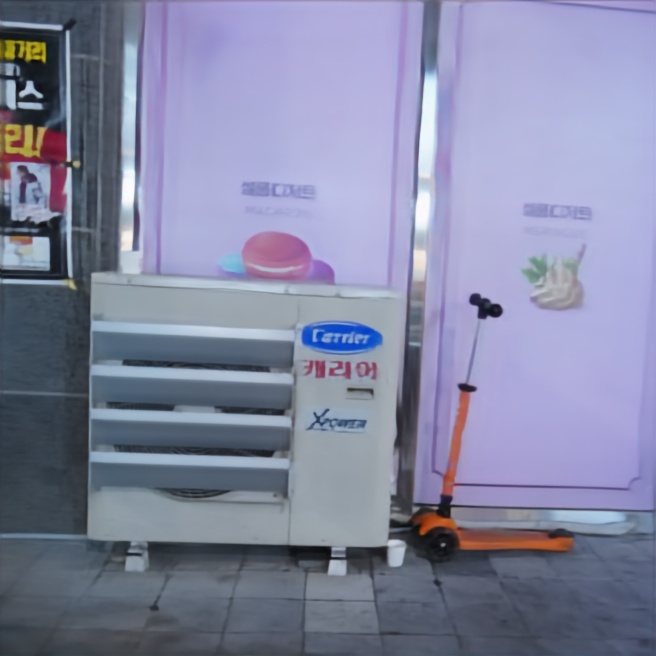}
            \caption{MIMOUNet~\cite{cho21rethinking}.}
        \end{subfigure} &
        \begin{subfigure}[t]{0.2\linewidth}
            \centering
            \includegraphics[width=\linewidth,trim=250 200 240 290, clip]{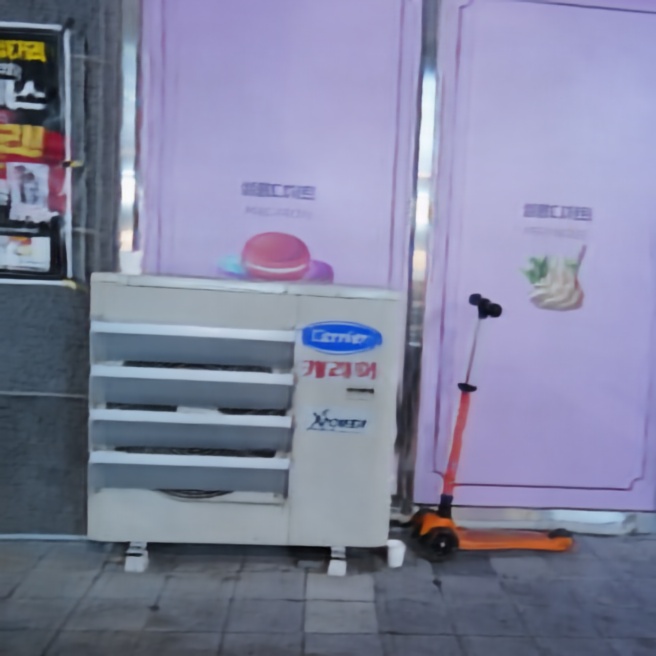}
            \caption{UNet-T (M, $16$).}
        \end{subfigure} &
        \begin{subfigure}[t]{0.2\linewidth}
            \centering
            \includegraphics[width=\linewidth,trim=250 200 240 290, clip]{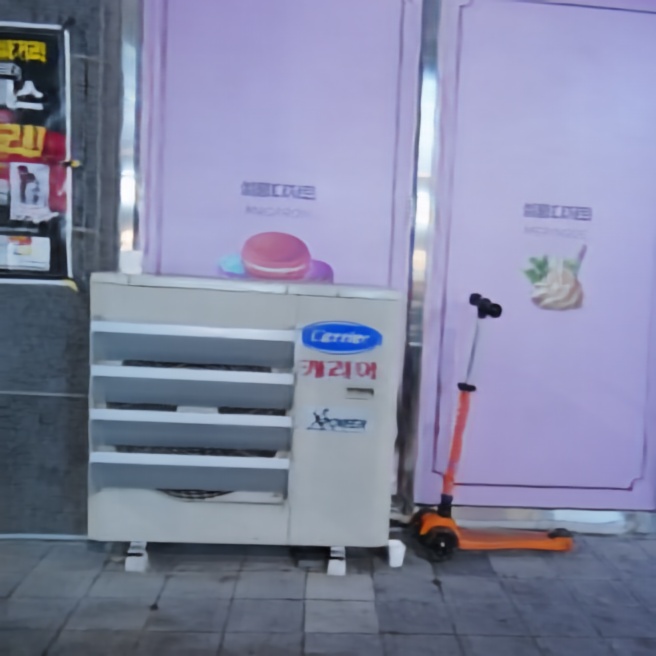}
            \caption{UNet-T (L, $16$).}
        \end{subfigure} &
        \begin{subfigure}[t]{0.2\linewidth}
            \centering
            \includegraphics[width=\linewidth,trim=250 200 240 290, clip]{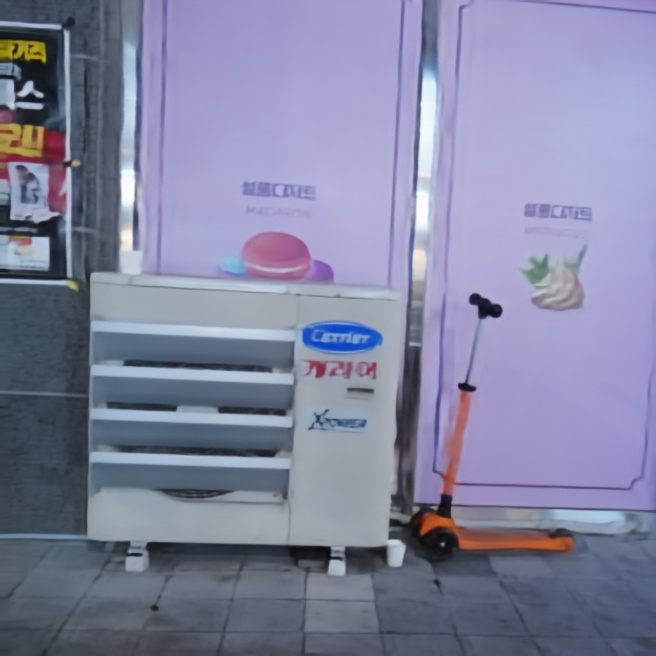}
            \caption{UNet-T (A, $16$).}
        \end{subfigure} \\
    \end{tabular} 
    }
    \caption{Camera shake removal from the Realblur-J test set~\cite{rim20dataset}. Please zoom on a computer screen.}
    \label{fig:shake_carrier}
\end{figure}

\begin{figure}[h!]
    \centering
    \resizebox{\linewidth}{!}{
    \begin{tabular}{ccccc}
        \begin{subfigure}[t]{0.2\linewidth}
            \centering
            \includegraphics[width=\linewidth,trim=50 350 300 170, clip]{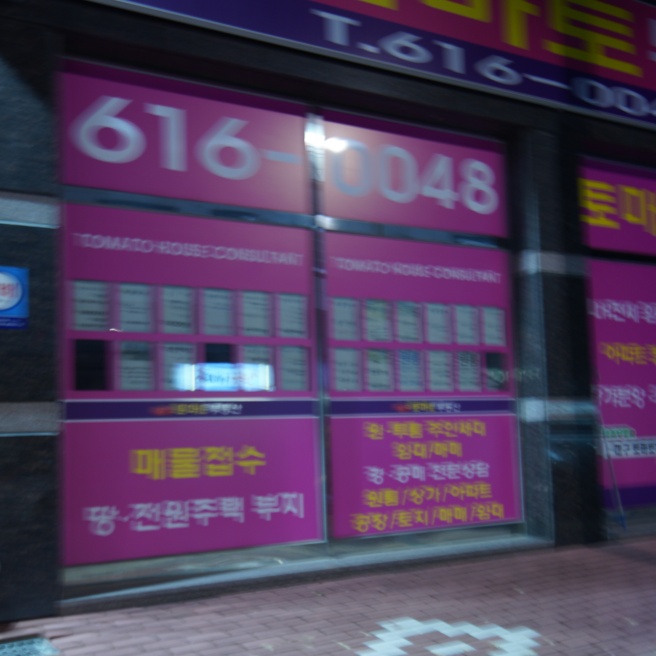}
            \caption{Blurry.}
        \end{subfigure} &
        \begin{subfigure}[t]{0.2\linewidth}
            \centering
            \includegraphics[width=\linewidth,trim=50 350 300 170, clip]{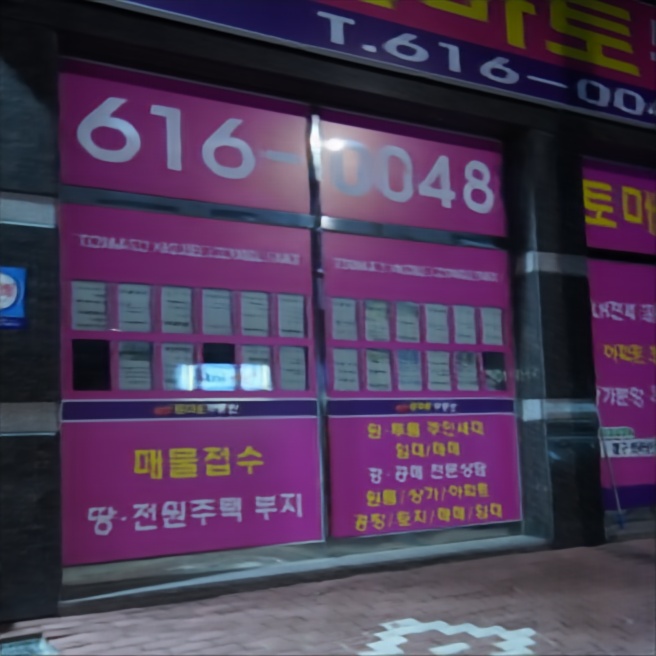}
            \caption{UNet-T.}
        \end{subfigure} &
        \begin{subfigure}[t]{0.2\linewidth}
            \centering
            \includegraphics[width=\linewidth,trim=50 350 300 170, clip]{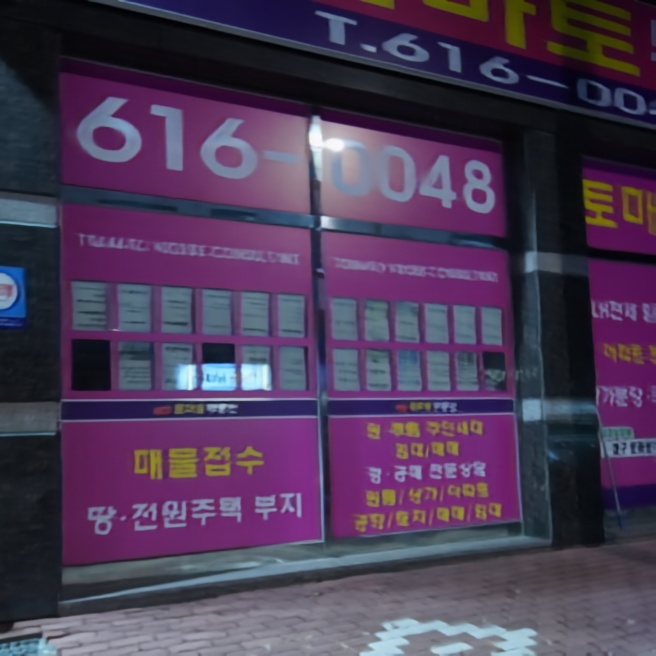}
            \caption{UNet-T (M, $4$).}
        \end{subfigure} &
        \begin{subfigure}[t]{0.2\linewidth}
            \centering
            \includegraphics[width=\linewidth,trim=50 350 300 170, clip]{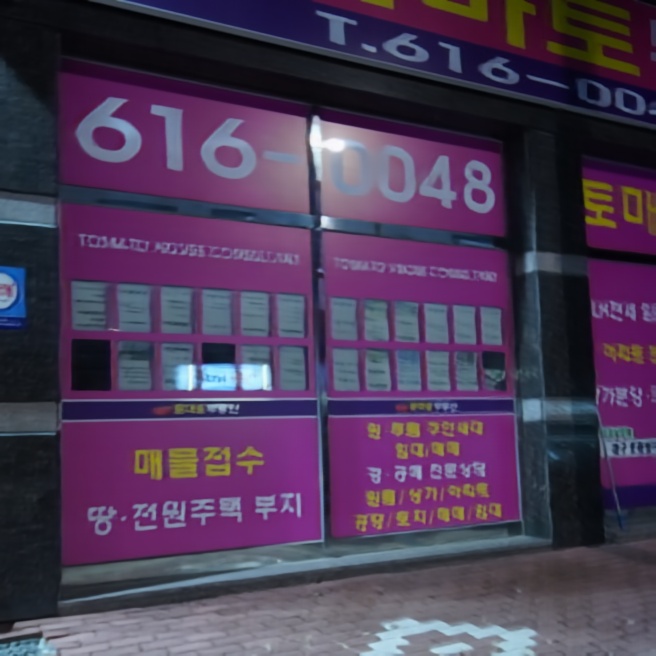}
            \caption{UNet-T (L, $4$).}
        \end{subfigure} &
        \begin{subfigure}[t]{0.2\linewidth}
            \centering
            \includegraphics[width=\linewidth,trim=50 350 300 170, clip]{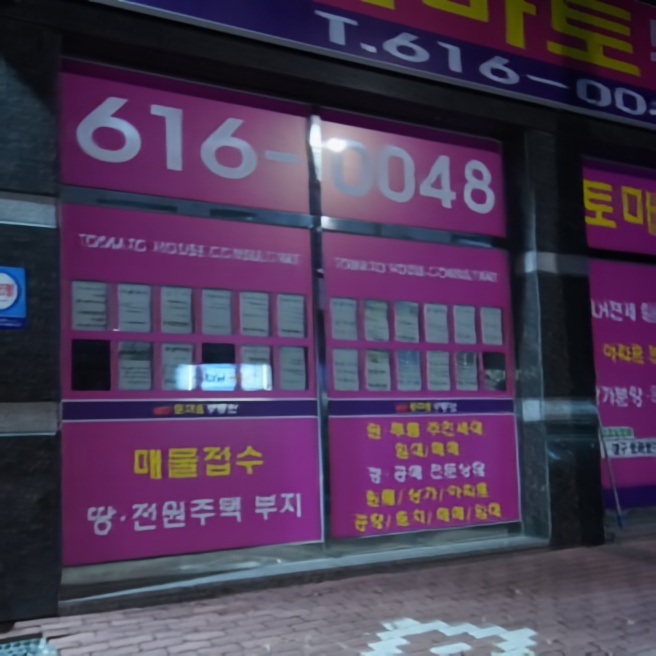}
            \caption{UNet-T (A, $4$).}
        \end{subfigure} \\
        \begin{subfigure}[t]{0.2\linewidth}
            \centering
            \includegraphics[width=\linewidth,trim=50 350 300 170, clip]{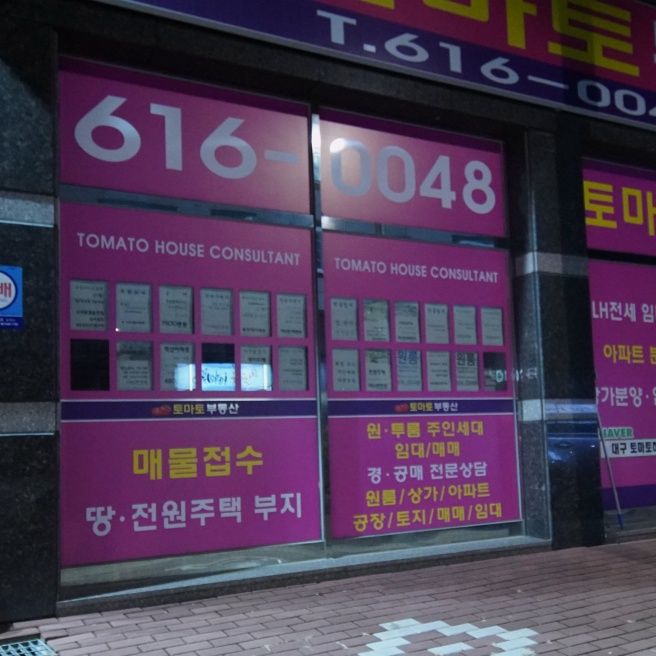}
            \caption{Target.}
        \end{subfigure} &
        \begin{subfigure}[t]{0.2\linewidth}
            \centering
            \includegraphics[width=\linewidth,trim=50 350 300 170, clip]{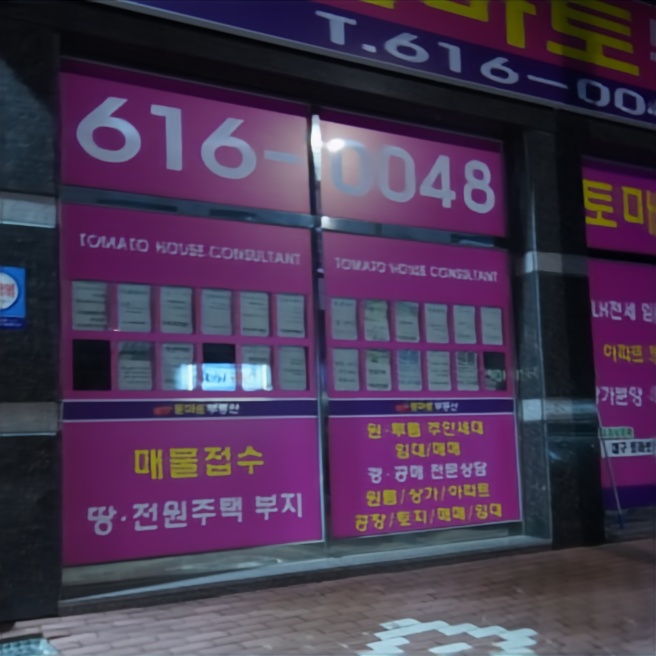}
            \caption{MIMOUNet~\cite{cho21rethinking}.}
        \end{subfigure} &
        \begin{subfigure}[t]{0.2\linewidth}
            \centering
            \includegraphics[width=\linewidth,trim=50 350 300 170, clip]{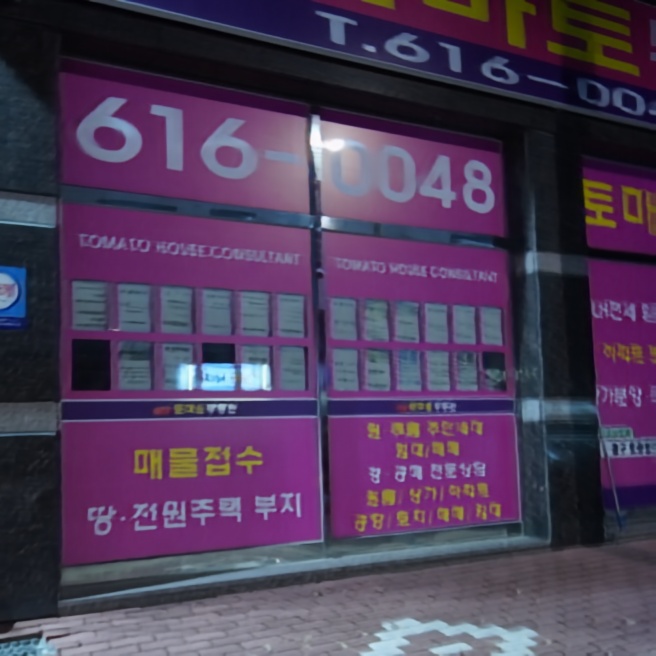}
            \caption{UNet-T (M, $16$).}
        \end{subfigure} &
        \begin{subfigure}[t]{0.2\linewidth}
            \centering
            \includegraphics[width=\linewidth,trim=50 350 300 170, clip]{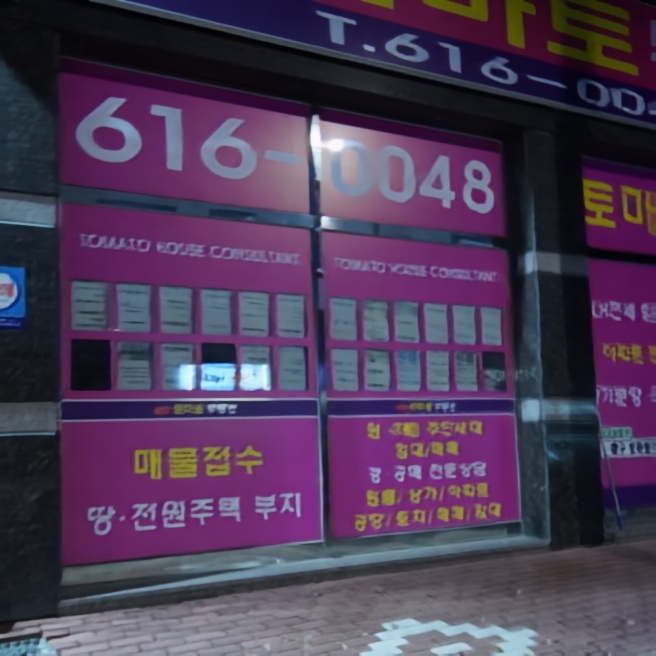}
            \caption{UNet-T (L, $16$).}
        \end{subfigure} &
        \begin{subfigure}[t]{0.2\linewidth}
            \centering
            \includegraphics[width=\linewidth,trim=50 350 300 170, clip]{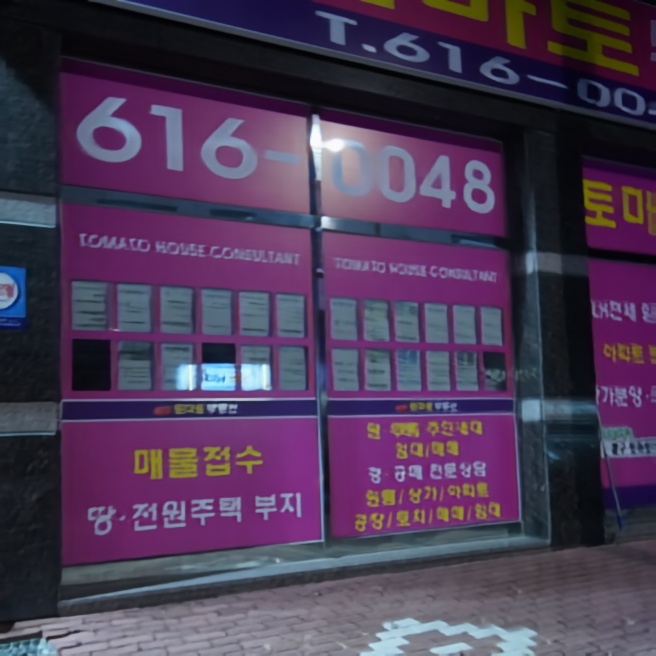}
            \caption{UNet-T (A, $16$).}
        \end{subfigure} \\
    \end{tabular} 
    }
    \caption{Camera shake removal from the Realblur-J test set~\cite{rim20dataset}. Please zoom on a computer screen.}
    \label{fig:shake_tomato}
\end{figure}

We provide additional crops from the Realblur-J dataset~\cite{rim20dataset}. We make $N$ crops collaborating
with the protocol described in the main paper: we slice the
image into $N$ overlapping patches with $25\%$ of overlap
to prevent fusion artifacts when building back the full-sized
image with the technique of \cite{schuler12blind}.
We show two results in Figures~\ref{fig:shake_carrier}
and \ref{fig:shake_tomato} where we illustrate the gradual positive impact
on a UNet-T to improve the pooling function $p$ between max pooling (M),
lambda layer (L) and self-attention (A), and two stack sizes $N$
in $\{4,16\}$. More refined strategies to predict a better global
feature $g$ leads to results comparable to MIMOUNet~\cite{cho21rethinking}
whereas the initial UNet-T hardly produces satisfactory results.

\subsection{Optical aberration}

\begin{figure}[h!]
    \centering
    \resizebox{\linewidth}{!}{
    \begin{tabular}{cccc}
        \begin{subfigure}[t]{0.25\linewidth}
            \centering
            \includegraphics[width=\linewidth,trim=300 0 0 220, clip]{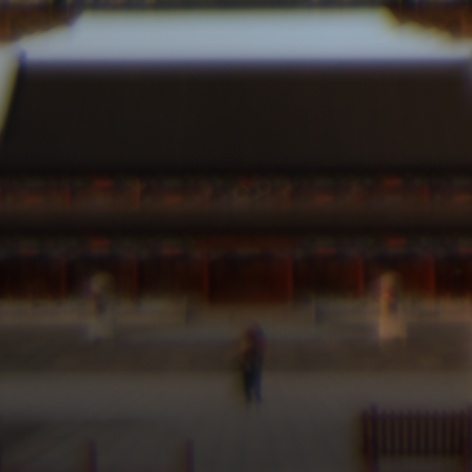}
            \caption{Aberrated.}
        \end{subfigure} &
        \begin{subfigure}[t]{0.25\linewidth}
            \centering
            \includegraphics[width=\linewidth,trim=300 0 0 220, clip]{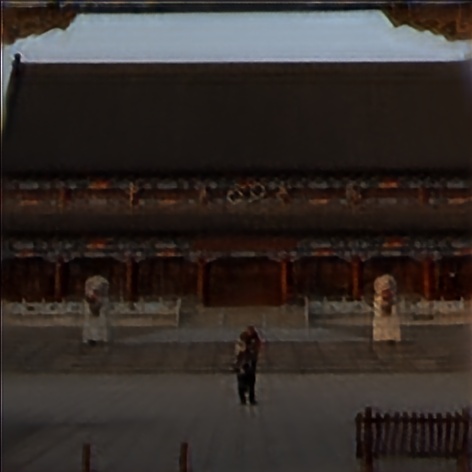}
            \caption{UNet-TX.}
        \end{subfigure} &
        \begin{subfigure}[t]{0.25\linewidth}
            \centering
            \includegraphics[width=\linewidth,trim=300 0 0 220, clip]{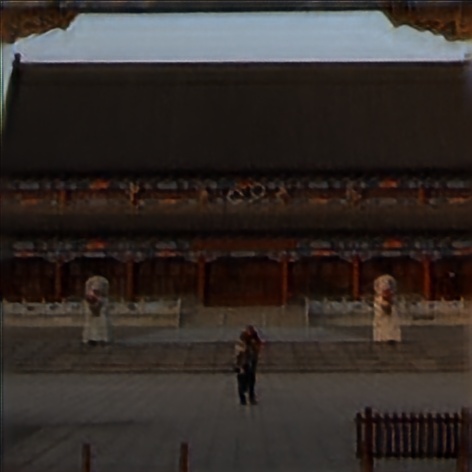}
            \caption{UNet-TX (M,4).}
        \end{subfigure} &
        \begin{subfigure}[t]{0.25\linewidth}
            \centering
            \includegraphics[width=\linewidth,trim=300 0 0 220, clip]{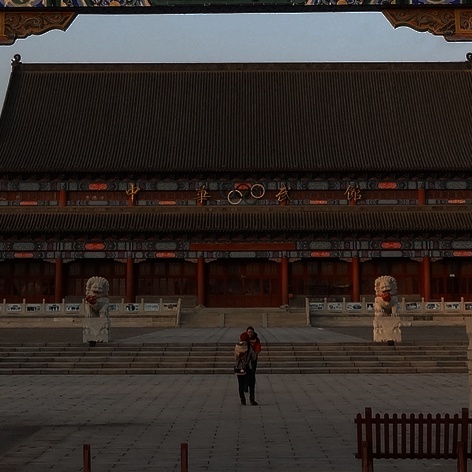}
            \caption{Target.}
        \end{subfigure} \\
    \end{tabular} 
    }
    \caption{Aberration removal example for the Canon 16-35mm f2.8L USM EI lens from~\cite{bauer18automatic}.}
    \label{fig:aberration_temple}
\end{figure}

\begin{figure}[h!]
    \centering
    \resizebox{\linewidth}{!}{
    \begin{tabular}{cccc}
        \begin{subfigure}[t]{0.25\linewidth}
            \centering
            \includegraphics[width=\linewidth,trim=250 20 70 300, clip]{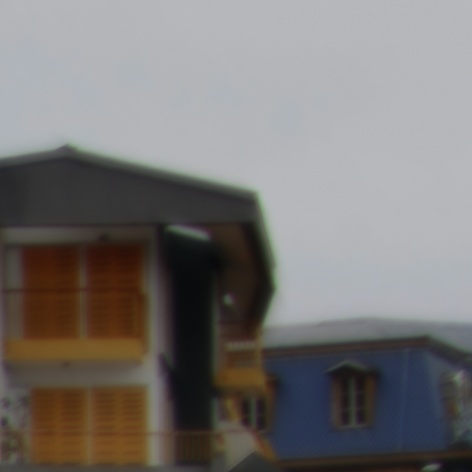}
            \caption{Aberrated.}
        \end{subfigure} &
        \begin{subfigure}[t]{0.25\linewidth}
            \centering
            \includegraphics[width=\linewidth,trim=250 20 70 300, clip]{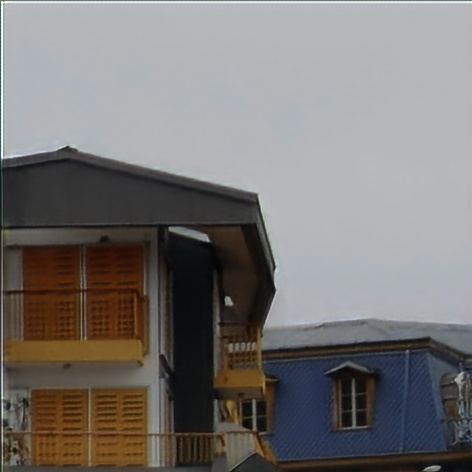}
            \caption{UNet-TX.}
        \end{subfigure} &
        \begin{subfigure}[t]{0.25\linewidth}
            \centering
            \includegraphics[width=\linewidth,trim=250 20 70 300, clip]{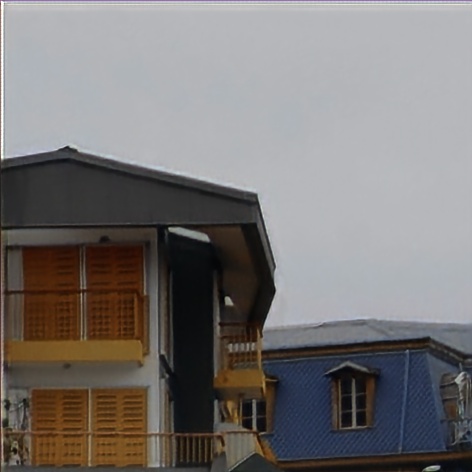}
            \caption{UNet-TX (M,4).}
        \end{subfigure} &
        \begin{subfigure}[t]{0.25\linewidth}
            \centering
            \includegraphics[width=\linewidth,trim=250 20 70 300, clip]{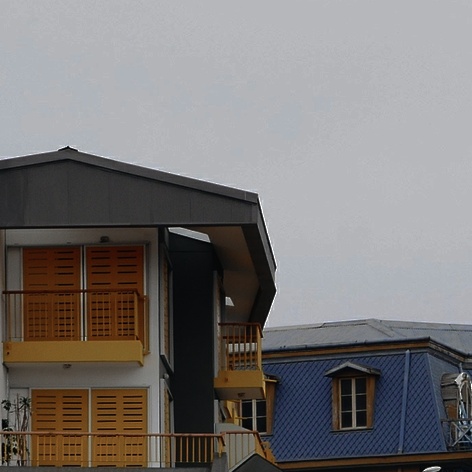}
            \caption{Target.}
        \end{subfigure} \\
    \end{tabular} 
    }
    \caption{Aberration removal example for the Canon 24mm f1.4L USM lens from~\cite{bauer18automatic}.}
    \label{fig:aberration_house}
\end{figure}

\begin{figure}[h!]
    \centering
    \resizebox{\linewidth}{!}{
    \begin{tabular}{cccc}
        \begin{subfigure}[t]{0.25\linewidth}
            \centering
            \includegraphics[width=\linewidth,trim=60 170 290 190, clip]{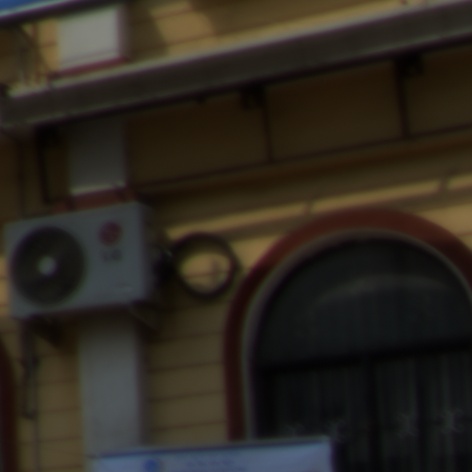}
            \caption{Aberrated.}
        \end{subfigure} &
        \begin{subfigure}[t]{0.25\linewidth}
            \centering
            \includegraphics[width=\linewidth,trim=60 170 290 190, clip]{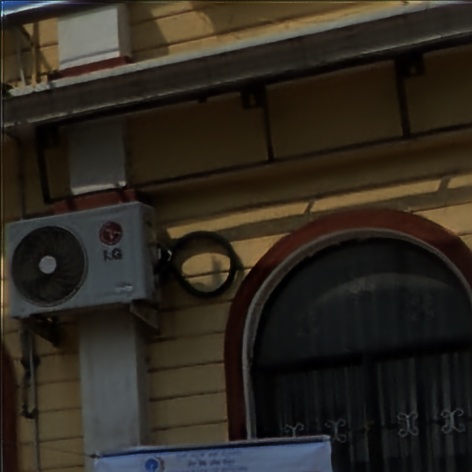}
            \caption{UNet-TX.}
        \end{subfigure} &
        \begin{subfigure}[t]{0.25\linewidth}
            \centering
            \includegraphics[width=\linewidth,trim=60 170 290 190, clip]{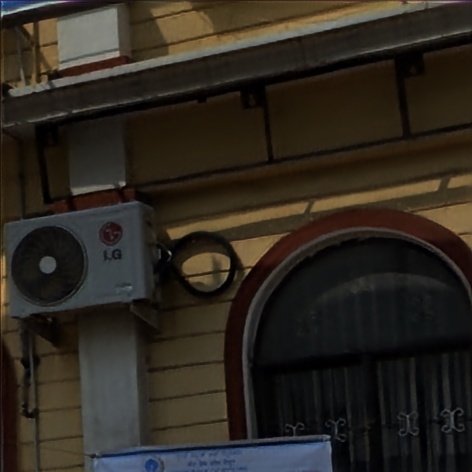}
            \caption{UNet-TX (M,4).}
        \end{subfigure} &
        \begin{subfigure}[t]{0.25\linewidth}
            \centering
            \includegraphics[width=\linewidth,trim=60 170 290 190, clip]{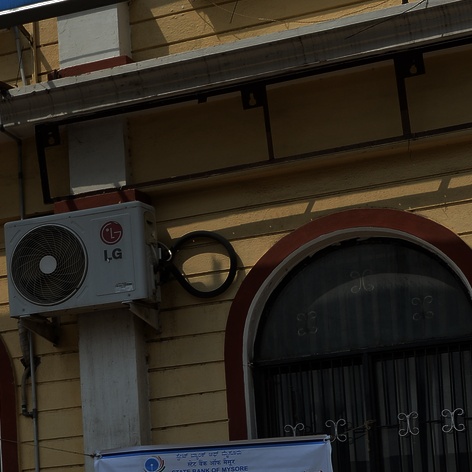}
            \caption{Target.}
        \end{subfigure} \\
    \end{tabular} 
    }
    \caption{Aberration removal example for the Canon 24mm f1.4L USM lens from~\cite{bauer18automatic}.}
    \label{fig:aberration_logo}
\end{figure}

We show qualitative examples synthesized with the 16mm and 24mm
lenses from \cite{bauer18automatic}.
We show results of remove of the aberrations in the corner in Figures~\ref{fig:aberration_temple},
\ref{fig:aberration_house} and \ref{fig:aberration_logo}.
Even small details and textures that seem lost because
of the aberrations may be salvaged with the collaborative scheme.

\end{document}